\documentclass[10pt,journal,compsoc]{IEEEtran}

\ifCLASSOPTIONcompsoc
  \usepackage[nocompress]{cite}
\else
  \usepackage{cite}
\fi
\ifCLASSINFOpdf
\else
\fi

\usepackage{titletoc}
\usepackage{booktabs} 
\usepackage{longtable} 
\usepackage{array}     
\usepackage{caption}   

\usepackage[utf8]{inputenc} 
\usepackage[T1]{fontenc}    
\usepackage{hyperref}       
\usepackage{url}            
\usepackage{booktabs}       
\usepackage{amsfonts}       
\usepackage{nicefrac}       
\usepackage{microtype}      

\usepackage{wrapfig}
\usepackage[utf8]{inputenc} 
\usepackage[T1]{fontenc}    
\usepackage{hyperref}       
\usepackage{url}            
\usepackage{booktabs}       
\usepackage{amsfonts}       
\usepackage{tcolorbox}   
\usepackage{nicefrac}       
\usepackage{microtype}      

\usepackage{amsthm}

\newtheorem{theorem}{Theorem}[section]

\usepackage{booktabs}
\usepackage{subfigure}
\usepackage{enumitem}
\usepackage{rotating}
 
\usepackage[utf8]{inputenc}
\usepackage{algorithm2e}
\usepackage{amsmath,amssymb,amsfonts}

\usepackage{colortbl}
\usepackage{multirow}
\usepackage{booktabs}
\usepackage{adjustbox}
\definecolor{grey}{rgb}{0.89,0.71,0.57}
\definecolor{pink}{rgb}{1,0.94,1}
\definecolor{purple}{rgb}{0.84,0.78,1}
\definecolor{white}{rgb}{1,1,1}
\usepackage{mathtools}
\usepackage[utf8]{inputenc}
\usepackage{hyperref}
\usepackage{hyperref}
\usepackage{amsmath}
\usepackage{bm}

\definecolor{mydarkblue}{rgb}{0,0.08,0.45}
\hypersetup{
    colorlinks=true,
    linkcolor=mydarkblue,
    citecolor=mydarkblue,
    filecolor=mydarkblue,
    urlcolor=mydarkblue,
    pdfview=FitH
}
\usepackage{tabularx} 
\usepackage{array}    
\usepackage[table]{xcolor} 
\usepackage{booktabs}
\usepackage{multirow}
\usepackage{graphicx} 
\usepackage{amsmath}  
\usepackage{amsfonts} 
\usepackage{fontawesome5} 

\definecolor{confICLR}{HTML}{D1E8FF} 
\definecolor{confNeurIPS}{HTML}{FFD9D9} 
\definecolor{confICML}{HTML}{D4EFDF} 
\definecolor{confKDD}{HTML}{FFF3CD} 
\definecolor{confMICCAI}{HTML}{E8DAEF} 
\definecolor{confCVPR}{HTML}{D0F0F0} 
\definecolor{confICCV}{HTML}{FFE0F1} 
\definecolor{confACMMM}{HTML}{E0E0A0} 

\newcommand{\ConfTag}[3]{\colorbox{#1}{\textcolor{black}{\scriptsize\textsc{#2 #3}}}}

\newcommand{\ICLR}[1]{\ConfTag{confICLR}{ICLR}{#1}}
\newcommand{\NeurIPS}[1]{\ConfTag{confNeurIPS}{NeurIPS}{#1}}
\newcommand{\NIPS}[1]{\ConfTag{confNeurIPS}{NIPS}{#1}}
\newcommand{\ICML}[1]{\ConfTag{confICML}{ICML}{#1}}
\newcommand{\KDD}[1]{\ConfTag{confKDD}{KDD}{#1}}
\newcommand{\MICCAI}[1]{\ConfTag{confMICCAI}{MICCAI}{#1}}
\newcommand{\CVPR}[1]{\ConfTag{confCVPR}{CVPR}{#1}}
\newcommand{\ICCV}[1]{\ConfTag{confICCV}{ICCV}{#1}}
\newcommand{\MM}[1]{\ConfTag{confACMMM}{MM}{#1}}

\usepackage{xcolor}
\definecolor{darkblue}{rgb}{0.0, 0.0, 0.55}
\definecolor{darkred}{rgb}{0.55, 0.0, 0.0}
\usepackage{wrapfig}
\usepackage{pifont}

\definecolor{cpink}{HTML}{FCCDE5}
\definecolor{cred}{HTML}{FFC2BA}
\definecolor{cyellow}{HTML}{FFFFB3}
\definecolor{cblue}{HTML}{B9DEFF}
\definecolor{cgreen}{HTML}{D7F3E7}
\definecolor{cneutral}{HTML}{CFCFCF}

\usepackage{xspace}
\newcommand{\ourmethod}{{\fontfamily{lmtt}\selectfont \textbf{Turb-L1}}\xspace}

\everypar{\looseness=-1}
\usepackage{titlesec}
\titlespacing{\paragraph}{%
  0pt}{
  0pt}{
  1em}
  
\usepackage[utf8]{inputenc} 
\usepackage[T1]{fontenc}    
\usepackage{hyperref}       
\usepackage{url}            
\usepackage{booktabs}       
\usepackage{amsfonts}       
\usepackage{nicefrac}       
\usepackage{microtype}      
\usepackage{xcolor}         
\newcommand{\insightbox}[1]{%
    \begin{tcolorbox}[
        colframe=black!70, 
        colback=blue!5,  
        boxrule=1pt, 
        arc=4mm,
        top=2mm,        
        bottom=2mm,      
        left=2mm        
        ]
        \includegraphics[width=0.32cm]{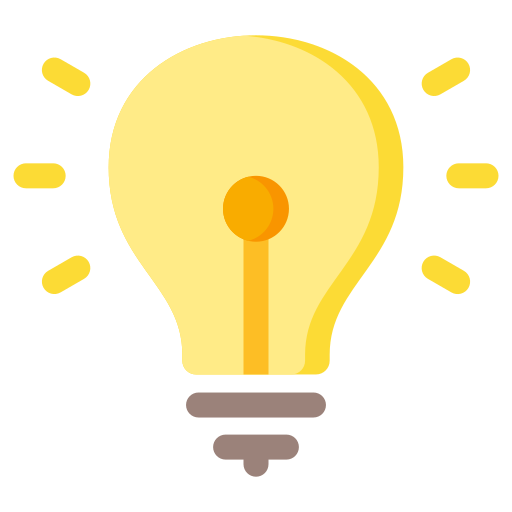}
        {\small#1}
    \end{tcolorbox}
}

\begin{document}
%
\title{\ourmethod: Achieving Long-term Turbulence Tracing By Tackling Spectral Bias}

\author{Hao Wu$^\dagger$ \quad Yuan Gao$^\dagger$ \quad Chang Liu  \quad Fan Xu \quad Fan Zhang \quad Zhihong Zhu \quad Yuqi Li  \quad Xian Wu \\ \quad Yuxuan Liang \quad Li Liu \quad Qingsong Wen   \quad Kun Wang$^*$ \quad Yu Zheng \quad Xiaomeng Huang$^*$ \\

\thanks{$^\dagger$These authors contributed equally to this work.}
\thanks{$^*$Corresponding authors: kun.wang@ntu.edu.sg, hxm@tsinghua.edu.cn}
\thanks{Hao Wu, Yuan Gao, Xiaomeng Huang are with Tsinghua University.}
\thanks{Chang Liu, Institute of Computing Technology, Chinese Academy of Sciences.}
\thanks{Fan Xu is with University of Science and Technology of China.}
\thanks{Yuxuan Liang is with Hong Kong University of Science and Technology (Guangzhou).}
\thanks{Fan Zhang is with The Chinese University of Hong Kong.}
\thanks{Zhihong Zhu, Xian Wu are with Tencent.}
\thanks{Yuqi Li is with The City College of New York.}
\thanks{Li Liu is with Chongqing University.}
\thanks{Qingsong Wen is with Squirrel AI learning.}
\thanks{Kun Wang is with Nanyang Technological University.}
\thanks{Yu Zheng is with JD iCity, JD Technology, Beijing 100176, China, and also with JD Intelligent Cities Research, Beijing, China.}

{\tt\footnotesize wuhao2022@mail.ustc.edu.cn} \quad \tt\footnotesize kun.wang@ntu.edu.sg \quad \tt\footnotesize hxm@tsinghua.edu.cn}

\markboth{IEEE TRANSACTIONS ON PATTERN ANALYSIS AND MACHINE INTELLIGENCE}
{Shell \MakeLowercase{\textit{et al.}}: Bare Demo of IEEEtran.cls for Computer Society Journals}
%



\IEEEtitleabstractindextext{%
\begin{abstract}

Accurately predicting the long-term evolution of turbulence is crucial for advancing scientific understanding and optimizing engineering applications. However, existing deep learning methods face significant bottlenecks in long-term autoregressive prediction, which exhibit excessive \textit{smoothing} and fail to accurately track complex fluid dynamics. Our extensive experimental and spectral analysis of prevailing methods provides an interpretable explanation for this shortcoming, identifying \textbf{Spectral Bias} as the core obstacle. Concretely, spectral bias is the inherent tendency of models to favor low-frequency, smooth features while overlooking critical high-frequency details during training, thus reducing fidelity and causing physical distortions in long-term predictions. Building on this insight, we propose \ourmethod{}, an innovative turbulence prediction method, which utilizes a \texttt{Hierarchical Dynamics Synthesis} mechanism within a multi-grid architecture to explicitly overcome spectral bias. It accurately captures cross-scale interactions and preserves the fidelity of high-frequency dynamics, enabling reliable long-term tracking of turbulence evolution. Extensive experiments on the 2D turbulence benchmark show that \ourmethod{} demonstrates excellent performance: \textbf{(I)} In long-term predictions, it reduces Mean Squared Error (MSE) by $80.3\%$ and increases Structural Similarity (SSIM) by over $9\times$ compared to the SOTA baseline, significantly improving prediction fidelity. \textbf{(II)} It effectively overcomes spectral bias, accurately reproducing the full enstrophy spectrum and maintaining physical realism in high-wavenumber regions, thus avoiding the spectral distortions or spurious energy accumulation seen in other methods. \textbf{(III)} It achieves outstanding long-term stability with a relative $L_2$ error as low as $0.444$, substantially outperforming baseline methods and proving its robustness. Our codes are available at~\url{https://github.com/Alexander-wu/TurbL1_AI4Science}
\end{abstract}

}

\maketitle

\IEEEdisplaynontitleabstractindextext

%
\IEEEpeerreviewmaketitle


\IEEEraisesectionheading{\section{Introduction}\label{sec:introduction}}

%
%
%
%


\IEEEPARstart{T}{urbulence}~\cite{sreenivasan1999fluid, davidson2015turbulence, duraisamy2019turbulence}, the intricate and chaotic dance of fluids shaping phenomena from weather systems to aerodynamic efficiency~\cite{bi2023accurate, lam2023learning, wu2024earthfarsser, gao2025oneforecast}. Accurate prediction of its long-term evolution is paramount not only for fundamental scientific understanding in fields like astrophysics~\cite{brandenburg2013astrophysical} and climate modeling~\cite{jackson2008parameterization,wyngaard2010turbulence} but also for critical engineering applications such as optimizing energy conversion~\cite{pope2001turbulent, borghi1988turbulent} and reducing drag~\cite{kim2003control, jimenez2018coherent}. However, its inherent multiscale nature~\cite{kolmogorov1995turbulence, kolmogorov1991local}, characterized by strong nonlinearity~\cite{ottino1989kinematics, monin1978nature}, chaotic dynamics, and the crucial energy cascade across scales, where fine structures dictate macroscopic behavior, renders long-term, high-fidelity prediction an enduring scientific frontier~\cite{moin1998direct, duraisamy2019turbulence}. 
\begin{figure}[h]
 \centering
 \includegraphics[width=0.48\textwidth]{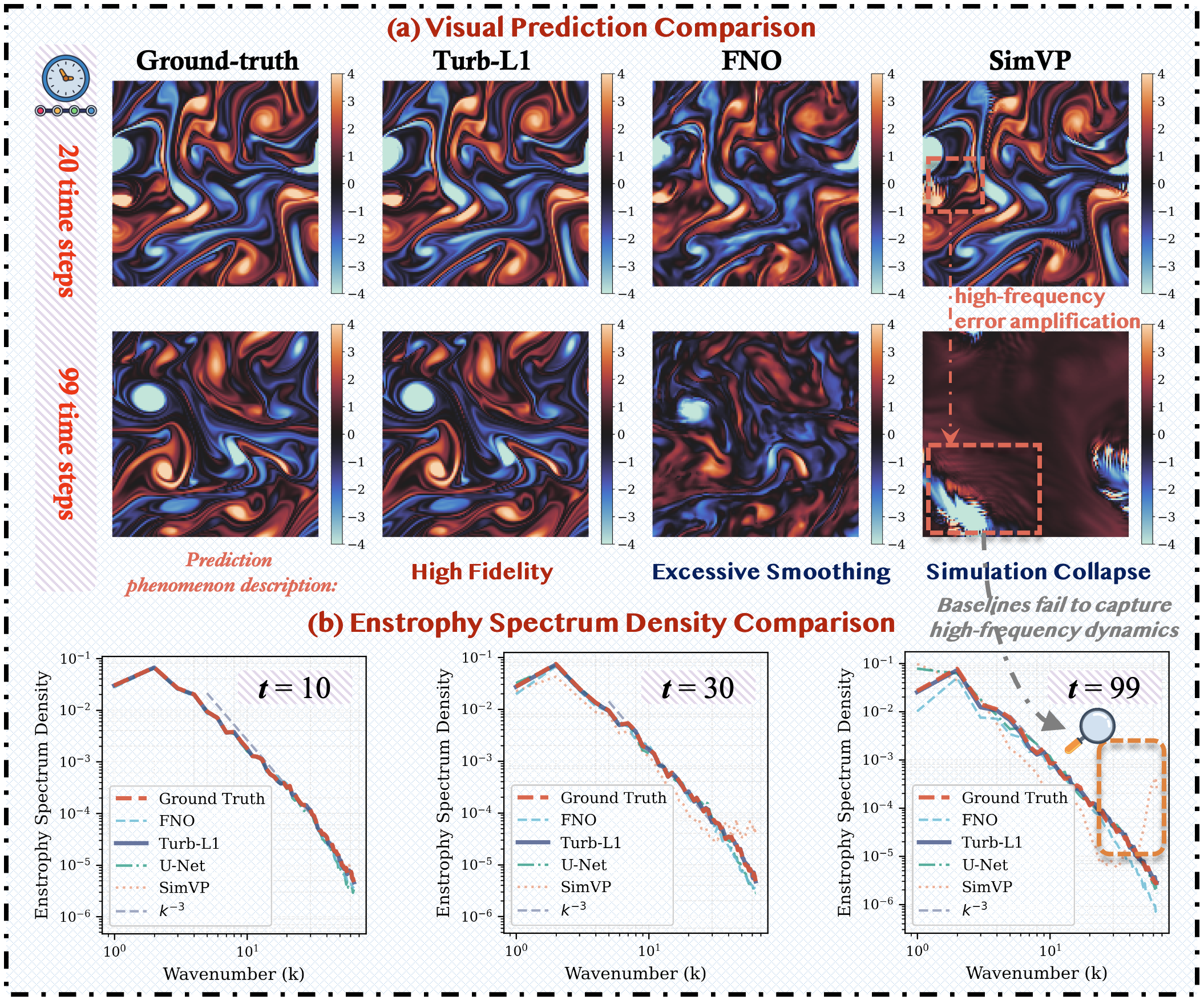}
 \caption{\textbf{Long-term turbulence prediction performance comparison.} \textbf{(a)} Visual comparison of vorticity fields ($t$=20, $t$=99). \ourmethod maintains high fidelity to the Ground-truth, while baselines exhibit excessive smoothing (e.g., FNO) or simulation collapse with artifacts (e.g., SimVP) driven by high-frequency error amplification. \textbf{(b)} Enstrophy spectrum density comparison ($t$=10, 30, 99). Baselines fail to capture high-frequency dynamics, increasingly deviating from the Ground Truth at high wavenumbers ($k$) over time, whereas \ourmethod accurately preserves spectral characteristics.}
 \label{intro:case}
\end{figure}

Traditional numerical simulations for fluid dynamics, such as direct numerical simulation (DNS)~\cite{moin1998direct, scardovelli1999direct,lee2015direct}, large eddy simulation (LES)~\cite{zhiyin2015large, piomelli1999large}, and Reynolds average Navier-Stokes (RANS) models~\cite{alfonsi2009reynolds,girimaji2006partially}, while foundational, grapple with a persistent trade-off between \textbf{\textit{computational cost}} and \textbf{\textit{physical accuracy}}, particularly hindering long-duration, high-resolution turbulence modeling. This challenge spurred interest in deep learning, which, despite successes in tasks like reconstruction~\cite{thapa2020dynamic, liu2020deep} and short-term forecasting~\cite{lopez2021model, li2021fourier, wu2024earthfarsser}, encounters a critical and widespread limitation when applied to the demanding problem of long-term autoregressive turbulence prediction using current mainstream architectures (e.g., Convolutional neural networks (CNNs)~\cite{he2019mgnet}, Graph neural networks (GNNs)~\cite{lam2023learning, gao2025oneforecast}, Transformers~\cite{wu2024transolver}, Neural Operators~\cite{li2020neural, li2021fourier, raonic2023convolutional, wu2024neural}). Inherent architectural characteristics, such as resolution loss in CNNs, difficulty capturing localized high-frequencies in Transformers, or explicit/implicit filtering in Operators often lead to a failure to preserve crucial fine-scale dynamics. Consequently, while initial predictions may seem plausible, extended rollouts rapidly diverge from physical reality, manifesting as excessive smoothing, loss of vital vortex structures, severe energy spectrum distortions, and unphysical artifacts, as shown in Figure.~\ref{intro:case}a, ultimately undermining their reliability for long-range forecasting needs that traditional methods also struggled to meet efficiently.

\textit{This study delves into the fundamental reasons underpinning this prevalent failure} across diverse deep learning architectures. We identify and argue that the core issue stems from an inherent property of many standard models: spectral bias \cite{rahaman2019spectral, fridovich2022spectral, shi2022measuring}. This bias manifests as a tendency during training to preferentially learn smooth, large-scale, low-frequency components~\cite{sitzmann2020implicit}, while struggling to accurately capture and maintain the less energetic, yet dynamically critical, high-frequency, small-scale structures~\cite{tancik2020fourier}. For turbulence, whose very essence lies in the intricate cross-scale energy cascade and the dynamics of fine-scale vortices governing dissipation, this inherent learning preference is particularly destructive~\cite{karniadakis2021physics}. It leads models to systematically neglect or misrepresent the behavior of these vital high-frequency components, disrupting the proper energy transfer across scales~\cite{bai2021predicting}. As a direct consequence, even small initial errors, particularly in the poorly learned high-frequency range, rapidly amplify during the autoregressive prediction process within the nonlinear system~\cite{pathak2022fourcastnet}. This amplification directly leads to the observed degradation of fine structures and emergence of artifacts seen visually (Figure.~\ref{intro:case}a), manifests as anomalous high-wavenumber energy behavior and severe spectral distortions (Figure.~\ref{intro:case}b), and ultimately hinders the ability to accurately trace the turbulence evolution over extended periods.

\insightbox{\textit{How then can we fundamentally improve prediction if inherent spectral bias prevents accurately tracing turbulence long-term? This work demonstrates a path forward by \textbf{tackling } what we identify as a \textbf{spectral bias}, thereby \textbf{achieving} reliable \textbf{long-term turbulence tracing}.}}

Addressing the limitations of existing deep learning methods often compromised by spectral bias in long-term turbulence forecasting, this paper presents \ourmethod{}. Crucially, the design of \ourmethod{} is directly motivated by our interpretable identification of spectral bias as the primary impediment to accurate long-term prediction. Our method stems from a more foundational principle: learning the intrinsic spectral dynamics across interacting scales. To this end, we introduce the \textbf{\texttt{Hierarchical Dynamics Synthesis}} mechanism within a multi-grid framework. This process first employs operators highly sensitive to local, high-frequency variations to capture fine-grained turbulent features; subsequently, it utilizes mechanisms adept at global context integration to model the evolution of large-scale structures and their coupling with finer scales. Through the careful orchestration of this two-stage synergistic spectral feature synthesis process, explicit refining of high-frequency components before their integration into the global context, \ourmethod{} effectively captures the complex cross-scale energy transfer fundamental to turbulence. As demonstrated visually and spectrally in Figure~\ref{intro:case}, unlike baseline methods that succumb to excessive smoothing or simulation collapse, our approach preserves high-fidelity details and maintains physical consistency over extended rollouts. This design enables robust and accurate long-range predictions crucial for scientific and engineering applications. We conducted extensive experiments on challenging turbulence benchmarks, where \ourmethod{} achieves consistent state-of-the-art (SOTA) performance, demonstrating significant gains in long-term fidelity and stability.

\textbf{Experimental Observations.} Comparative experiments on the McWilliams 2D turbulence benchmark~\cite{mcwilliams1984emergence} validate the effectiveness of \ourmethod{}. \ding{182} \textbf{Delivers state-of-the-art long-term accuracy and fidelity:} At $t=99$, \ourmethod{} reduces MSE to $0.617$, an $80.3\%$ improvement over the best baseline FNO, and maintains SSIM at $0.682$, over $9\times$ higher than baselines like FNO, SimVP, and UNet, successfully preserving fine-scale structures. \ding{183} \textbf{Overcomes spectral bias, maintaining physical realism:} As shown in Figure~\ref{intro:case}b, it accurately captures the full enstrophy spectrum, crucially preserving high-frequency dynamics. This avoids issues like the excessive smoothing seen in FNO or the error explosion leading to simulation collapse observed in SimVP, where MSE increased over $14\times$ from $t=20$ to $t=99$. \ding{184} \textbf{Exhibits superior long-term stability:} At $t=99$, the relative L2 error is only $0.444$. This is at least $57\%$ lower than baselines typically exceeding $1.0$, such as FNO at $1.041$ and SimVP at $1.321$. This ensures reliable tracking of turbulence dynamics, highlighting its significant potential for long-term scientific and engineering prediction tasks.

To the best of our knowledge, \ourmethod{} represents the first systematic study addressing long-term modeling in turbulent scenarios. We summarize our key contributions as follows: \ding{224} We establish spectral bias as a critical, interpretable factor limiting current deep learning approaches in long-term turbulence predictions. \ding{224} We introduce \ourmethod{}, a novel method designed to explicitly counteract this bias, thereby enabling unprecedented long-term prediction fidelity and physical realism. \ding{224} Comprehensive validation demonstrating \ourmethod{}'s state-of-the-art long-term prediction capabilities.





\section{Related Work}
\label{sec:related_work}

\subsection{Traditional Numerical Simulation of Turbulence}
Traditional methods for turbulence simulation, such as Direct Numerical Simulation (DNS)~\cite{moin1998direct, scardovelli1999direct}, Large Eddy Simulation (LES)~\cite{zhiyin2015large, piomelli1999large}, and Reynolds-Averaged Navier-Stokes (RANS)~\cite{alfonsi2009reynolds, girimaji2006partially}, serve as the foundation of fluid dynamics research. While DNS provides the highest fidelity by resolving all spatio-temporal scales, its prohibitive computational cost makes it impractical for long-term simulations at high Reynolds numbers common in engineering applications~\cite{lee2015direct}. LES and RANS reduce this cost by modeling sub-grid scales or averaging turbulent fluctuations, respectively. However, these approaches rely on closure models that often lack universality~\cite{maulik2017neural} and can introduce significant errors in complex flow regimes~\cite{ling2016reynolds, duraisamy2019turbulence, kochkov2021machine}. These methods present a fundamental trade-off between computational efficiency and physical accuracy. In contrast to these methods, which depend on hand-crafted models, our work aims to learn the complete evolution operator directly from high-fidelity data, seeking to achieve near-DNS accuracy in long-term predictions at a fraction of the computational cost.

\subsection{Deep Learning for Physical System Modeling}
Deep learning has recently shown immense potential for modeling physical systems. Models based on Convolutional Neural Networks (CNNs), such as U-Net~\cite{ronneberger2015u} and ResNet~\cite{he2016deep}, excel at capturing local spatial features but are prone to losing high-frequency information through repeated pooling and convolution operations, leading to over-smoothing in long-term autoregressive predictions~\cite{sanchez2020learning, wu2024pure}. The Fourier Neural Operator (FNO)~\cite{li2021fourier, kovachki2023neural} learns PDE solution operators efficiently by performing global convolutions in the frequency domain; however, its nature as a low-pass filter systematically neglects small-scale dynamics. Transformer architectures~\cite{gao2022earthformer, lam2023learning, wu2024transolver, wu2025advanced} capture global dependencies but are not inherently sensitive to local, high-frequency structural changes. Prevailing deep learning approaches, including CNNs, Transformers, and spatiotemporal models like SimVP~\cite{tan2022simvp}, universally struggle with either over-smoothing or simulation collapse in long-term turbulence forecasting. This paper systematically argues that the common root of these failures is the inherent \textbf{spectral bias}~\cite{rahaman2019spectral, kiessling2022computable, fang2024addressing} of neural networks.

\subsection{Architectures for Addressing Spectral Bias}
Spectral bias, the tendency of neural networks to preferentially learn low-frequency, smooth functions~\cite{rahaman2019spectral}, is a key reason for the loss of high-frequency details. Although prior work has leveraged multi-grid concepts (e.g., MGNet~\cite{he2019mgnet}) to improve network performance or attempted to implicitly introduce high-frequency information through techniques like positional encodings, these approaches do not fundamentally resolve the issue~\cite{mildenhall2021nerf}. Our model, \ourmethod{}, introduces a novel and explicit solution. It not only employs a multi-grid framework for scale separation but, more critically, incorporates a \textbf{Hierarchical Dynamics Synthesis (HDS)} mechanism. This mechanism forces the model to learn and preserve full-spectrum dynamics by routing features through specialized pathways—using CNNs for high frequencies~\cite{dai2021coatnet, wu2021cvt,guo2022cmt} and Transformer-like operators~\cite{dosovitskiy2020image, wu2024spatio} for low frequencies. This architecture-driven strategy, designed to directly counteract spectral bias, is the core distinction between \ourmethod{} and existing methods and is the key to its unprecedented performance in long-term, high-fidelity turbulence tracing.
\section{Method}
\begin{figure*}[t]
    \centering
    \includegraphics[width=\textwidth]{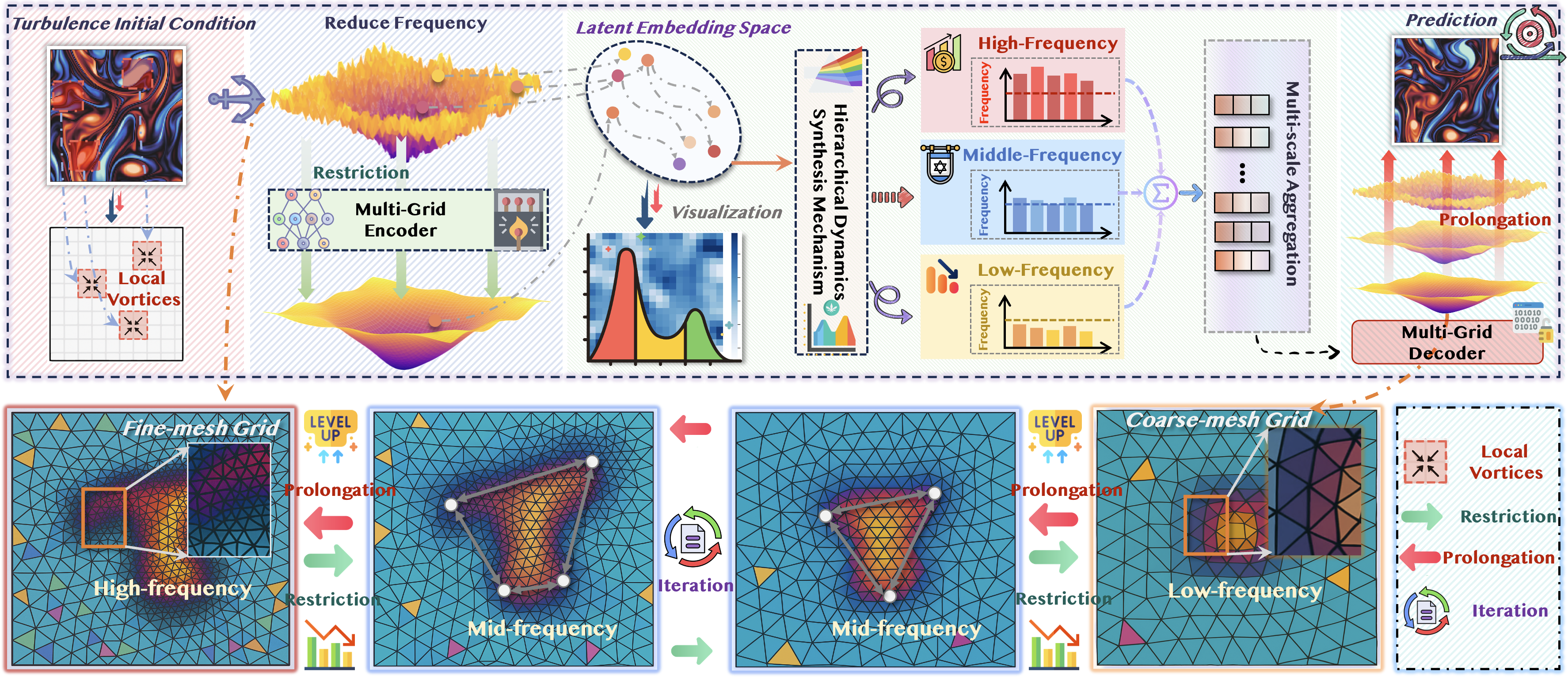} 
    \caption{
    Overview of the \ourmethod{} architecture.
    The model initially downscales the turbulence initial condition and embeds it into a latent space via a multi-grid encoder ($\mathcal{E}$).
    The core \textbf{Hierarchical Dynamics Synthesis Mechanism} ($\mathcal{HDS}$) then explicitly synthesizes dynamical features across different frequencies (high $\omega_H$, middle $\omega_M$, and low $\omega_L$) within this latent space.
    Subsequently, multi-scale aggregated features are upscaled by a multi-grid decoder ($\mathcal{D}$), ultimately generating high-fidelity turbulence predictions.
    The lower panel details the multi-grid operations---Restriction ($\mathcal{R}$), Prolongation ($\mathcal{P}$), and Iteration ($\circlearrowleft$)---illustrating how the model processes and transfers information across various grid scales to effectively capture cross-scale interactions and overcome spectral bias.
}
\label{fig:architecture}
\end{figure*}

\subsection{Problem Definition}
Forecasting the spatiotemporal evolution of turbulence, a complex fluid dynamic phenomenon governed by the Navier-Stokes equations and often characterized by its vorticity field $\omega(\mathbf{x}, t)$ in 2D incompressible flows, remains a significant challenge. The dynamics of $\omega(\mathbf{x}, t)$ are described by the vorticity transport equation, often written in conservative form for incompressible flow ($\nabla \cdot \mathbf{u} = 0$):
\begin{equation}
    \frac{\partial \omega(\mathbf{x},t)}{\partial t} + \nabla \cdot (\mathbf{u}(\mathbf{x},t)\omega(\mathbf{x},t)) = \nu \nabla^2 \omega(\mathbf{x},t) + f(\mathbf{x},t),
    \label{eq:vorticity_transport_formal}
\end{equation}
where $\mathbf{u}(\mathbf{x},t)$ is the velocity field, $\nu$ denotes kinematic viscosity, and $f(\mathbf{x},t)$ represents external forcing. Our objective is to develop a deep learning surrogate model, parameterized by $\theta$, denoted as $\mathcal{F}(\cdot; \theta)$, which approximates the solution operator for long-term autoregressive forecasting. Given an initial state $\omega_0(\mathbf{x}) = \omega(\mathbf{x}, t_0)$, the model iteratively predicts future vorticity states:
\begin{equation}
    \hat{\omega}(\mathbf{x}, t_{k+1}) = \mathcal{F}(\hat{\omega}(\mathbf{x}, t_k); \theta),
    \label{eq:autoregressive_prediction_formal}
\end{equation}
where $t_{k+1} = t_k + \Delta t$. The primary challenge is maintaining long-term high-fidelity and physical realism, especially in capturing crucial multi-scale interactions and high-frequency details often compromised by spectral bias in current learning architectures.

\subsection{Overview of the Proposed Method: \ourmethod{}}
To approximate the turbulence solution operator $\mathcal{F}(\cdot; \theta)$ (Eq.~\ref{eq:autoregressive_prediction_formal}) and mitigate spectral bias for long-term high-fidelity forecasting, we introduce \ourmethod{}, as shown in Figure~\ref{fig:architecture}. This framework transforms an input vorticity field $\omega(\mathbf{x}, t_k)$ (or a history thereof) into a predicted state $\hat{\omega}(\mathbf{x}, t_{k+1})$ through a meticulously designed sequence of learnable operators. \ourmethod{} first employs a multi-grid encoder $\mathcal{E}$ to extract hierarchical latent features, followed by a core Hierarchical Dynamics Synthesis mechanism $\mathcal{HDS}$ that evolves these features while preserving cross-spectral fidelity, and finally a multi-grid decoder $\mathcal{D}$ to reconstruct the output. This composite mapping, critical for robust autoregressive prediction, is denoted as:
\begin{equation}
    \hat{\omega}(\mathbf{x}, t_{k+1}) = (\mathcal{D} \circ \mathcal{HDS} \circ \mathcal{E})(\omega(\mathbf{x}, t_k); \theta) \equiv \mathcal{F}(\omega(\mathbf{x}, t_k); \theta),
    \label{eq:overall_architecture_ultra_concise}
\end{equation}
where $\theta$ are all learnable parameters $\{\theta_{\mathcal{E}}, \theta_{\mathcal{HDS}}, \theta_{\mathcal{D}}\}$, and for $k>0$, $\omega(\mathbf{x}, t_k)$ becomes $\hat{\omega}(\mathbf{x}, t_k)$.

\subsection{Projecting Turbulence onto a Latent Manifold Space}
The initial stage of \ourmethod{} involves the Multi-Grid Encoder, $\mathcal{E}(\cdot; \theta_{\mathcal{E}})$, transforming the high-dimensional input vorticity field $\omega(\mathbf{x}, t_k)$ into a structured, multi-scale latent representation. Turbulent flows, rich in high-frequency, small-scale structures, pose challenges for direct processing.

Inspired by multi-grid methods, our encoder $\mathcal{E}$ hierarchically projects data onto a latent manifold $\mathcal{M}_{\omega}$ where dynamics are assumed to be smoother. This involves restriction operators $\mathcal{R}_l$ and learned transformations $\phi_l(\cdot; \theta_{\mathcal{E}}^{(l)})$. The restriction operator $\mathcal{R}_l$, as shown in Figure~\ref{fig:architecture}, projects features to a coarser grid. \textbf{In our implementation}, each restriction operator is a module comprising a $3 \times 3$ strided convolution, followed by \texttt{GroupNorm} and a \texttt{LeakyReLU} activation. The strides are strategically chosen from the sequence `[1, 2, 1, 2, ...]` to ensure a gradual downsampling process. This crucial step reduces dimensionality and acts as a localized filter, preferentially passing smoother, large-scale signal components to coarser levels. The learned transformations $\phi_l$ then process these restricted features:
\begin{equation}
    \mathbf{z}^{(0)}(\mathbf{x}, t_k) = \phi_0(\omega(\mathbf{x}, t_k); \theta_{\mathcal{E}}^{(0)})
\end{equation}
\begin{equation}
\begin{aligned}
       \mathbf{z}^{(l+1)}(\mathbf{x}', t_k) &= \phi_{l+1}(\mathcal{R}_l(\mathbf{z}^{(l)}(\mathbf{x}, t_k)); \theta_{\mathcal{E}}^{(l+1)}), \\ & \text{for} ~~ l = 0, \dots, L-1
    \label{eq:encoder_mg_projection_concise_fig_ref}
\end{aligned}
\end{equation}
where $\mathbf{z}^{(l)}$ is the latent representation at grid level $l$. This multi-grid encoding, backed by our specific convolutional design, hierarchically disentangles features for effective cross-scale modeling, enhances computational efficiency, and facilitates dynamics learning on smoother latent manifolds by mitigating the stiffness and high dimensionality of the original PDE solution space, thus providing a rich, multi-scale context for the HDS mechanism.

\subsection{Hierarchical Dynamics Synthesis for Full-Spectrum Fidelity}
\label{sec:hds_corrected}

The Hierarchical Dynamics Synthesis (HDS) mechanism, $\mathcal{HDS}(\cdot; \theta_{\mathcal{HDS}})$, is pivotal to \ourmethod{}'s strategy for mitigating spectral bias. It operates on the latent representation $\mathbf{z}_{\text{enc}} \in \mathcal{M}_{\omega}$ from the multi-grid encoder $\mathcal{E}$, precisely evolving turbulence dynamics within this manifold. The HDS design leverages insights into the distinct frequency responses of neural operators: Convolutional Neural Networks (CNNs) excel at capturing high-frequency local details, whereas Transformer-like architectures are adept at integrating low-frequency global information~\cite{park2022vision, schiaffini2024neural}. \ourmethod{} architecturally synergizes these complementary properties for precise, high-fidelity, full-spectrum turbulence evolution.

Models reliant on a single operator type often exhibit spectral bias, prioritizing high-energy, low-frequency components over crucial high-frequency fine structures in turbulence. To counter this, HDS employs a heterogeneous, multi-path dynamics feature synthesis framework (Figure~\ref{fig:architecture}, center), departing from homogeneous evolution. This framework concurrently processes $\mathbf{z}_{\text{enc}}$ to extract and evolve features across distinct frequency scales. For each band $\omega_s$ ($s \in \{H, M, L\}$ for high, middle, low), a tailored non-linear transformation $\mathcal{T}_{\omega_s}(\cdot; \theta_{\mathcal{HDS}}^{(s)})$ generates dynamic primitives $\mathbf{h}_{\omega_s}$:
\begin{equation}
    \mathbf{h}_{\omega_s} = \mathcal{T}_{\omega_s}(\mathbf{z}_{\text{enc}}; \theta_{\mathcal{HDS}}^{(s)}).
    \label{eq:hds_frequency_primitives_corrected_en}
\end{equation}
The specific architecture of each $\mathcal{T}_{\omega_s}$ is optimized for its designated frequency band.

For synthesizing \textit{high-frequency dynamic primitives $\mathbf{h}_{\omega_H}$}, $\mathcal{T}_{\omega_H}$ employs operators highly sensitive to local variations. Convolutional operations, particularly with small receptive field kernels $\kappa_j^{(H)}$, inherently function as high-pass filters. Consequently, $\mathcal{T}_{\omega_H}$ is instantiated via such convolutional architectures:
\begin{equation} \footnotesize
    [\mathcal{T}_{\omega_H}(\mathbf{z}_{\text{enc}})](\mathbf{x}') = \sigma \left( \sum_j \int_{\mathcal{N}(\mathbf{x}')} \kappa_j^{(H)}(\mathbf{x}' - \mathbf{y}') \mathbf{z}_{\text{enc}}(\mathbf{y}') d\mathbf{y}' + \mathbf{b}_j^{(H)} \right),
    \label{eq:high_freq_synthesis_conv_corrected_en}
\end{equation}
where $\sigma$ is an activation and $\mathbf{b}_j^{(H)}$ a bias. \textbf{Our implementation}, termed \texttt{HighFrequencyOperator}, realizes this with a residual block incorporating a $3 \times 3$ depth-wise convolutional positional encoding, a spatial attention mimic composed of a $5 \times 5$ depth-wise convolution flanked by $1 \times 1$ convolutions, and an MLP block built from $1 \times 1$ convolutions.

Conversely, synthesizing \textit{low-frequency dynamic primitives $\mathbf{h}_{\omega_L}$} is achieved by $\mathcal{T}_{\omega_L}$, which targets large-scale structures. The self-attention mechanism (MSA) in Transformers inherently acts as a low-pass filter. $\mathcal{T}_{\omega_L}$ is thus instantiable as a Global Adaptive Weighted Aggregation (GAWA) mechanism, rooted in MSA:
\begin{equation}
    [\mathcal{T}_{\omega_L}(\mathbf{z}_{\text{enc}})]_i = \sum_{k=1}^{N_h} \mathbf{W}_k^{(O)} \left( \sum_{j=1}^{N_T} \alpha_{ij}^{(k)} (\mathbf{W}_k^{(V)} \mathbf{z}_j) \right),
    \label{eq:low_freq_synthesis_gawa_corrected_en}
\end{equation}
where $\alpha_{ij}^{(k)} = \text{softmax}_j \left( \frac{(\mathbf{W}_k^{(Q)} \mathbf{z}_i)^T (\mathbf{W}_k^{(K)} \mathbf{z}_j)}{\sqrt{d_k}} \right)$, $N_T$ is the token count, and $\mathbf{W}_k^{(\cdot)}$ are learnable head projection matrices. \textbf{This is implemented as \texttt{LowFrequencyOperator}}, a Transformer-style block with standard multi-head self-attention and a linear MLP. Crucially, its residual connections are scaled by learnable parameters initialized to a small value ($10^{-6}$), akin to LayerScale, for stable training.

The concurrently generated dynamic primitives $\{\mathbf{h}_{\omega_H}, \mathbf{h}_{\omega_M}, \mathbf{h}_{\omega_L}\}$ are subsequently fused by a Multi-scale Dynamic Aggregation (MDA) module, $\mathcal{A}_{\text{MDA}}(\cdot, \cdot, \cdot; \theta_{\mathcal{HDS}}^{(\text{agg})})$. This process yields a unified latent representation $\mathbf{z}_{\text{HDS\_out}}$:
\begin{equation}
    \mathbf{z}_{\text{HDS\_out}} = \mathcal{A}_{\text{MDA}}(\mathbf{h}_{\omega_H}, \mathbf{h}_{\omega_M}, \mathbf{h}_{\omega_L}; \theta_{\mathcal{HDS}}^{(\text{agg})}).
    \label{eq:hds_aggregation_mda_corrected_en}
\end{equation}
This HDS strategy, with its dedicated processing and intelligent aggregation, forces the model to learn and preserve critical high-frequency details, fundamentally mitigating spectral bias.

\subsection{Mapping to Physical Space}
The Multi-Grid Decoder $\mathcal{D}$ receives the latent space features $\mathbf{z}_{\text{HDS\_out}}$. It reconstructs the turbulent vorticity field $\hat{\omega}(\mathbf{x}, t_{k+1})$ through a sequence of hierarchical operations. It iteratively applies a Prolongation Operator $\mathcal{P}_l$ (upsampling) followed by a learnable Feature Refinement Module $\psi_l(\cdot; \theta_{\mathcal{D}}^{(l)})$. \textbf{In our model}, the Prolongation operator is realized using a $3 \times 3$ transposed convolution (\texttt{ConvTranspose2d}), again followed by \texttt{GroupNorm} and \texttt{LeakyReLU}. A key feature is a skip-connection that concatenates the upsampled features with the corresponding high-resolution feature map saved from the encoder. This re-introduces fine-scale details lost during downsampling. Finally, an output transformation layer $\mathcal{T}_{\text{out}}$, implemented as a $1 \times 1$ convolution, maps the refined features to the final vorticity field. The overall mathematical formulation is:
\begin{equation}\small
\begin{aligned}
& \hat{\omega}(\mathbf{x}, t_{k+1}) = \mathcal{T}_{\text{out}} \\ & \left( \psi_0 \left( \mathcal{P}_0\left( \dots \psi_{L-1}\left(\mathcal{P}_{L-1} \left(\mathbf{z}_{\text{HDS\_out}}\right); \theta_{\mathcal{D}}^{(L-1)}\right) \dots \right); \theta_{\mathcal{D}}^{(0)}\right); \theta_{\mathcal{D}}^{(\text{out})} \right)
\end{aligned}
\end{equation}
where $\mathbf{z}_{\text{HDS\_out}}$ is the output from the HDS module, $\mathcal{P}_l$ denotes the Prolongation operator at stage $l$, and $\psi_l$ is the Feature Refinement Module.

\subsection{Theoretical Analysis}
We theoretically analyze the link between a model's spectral accuracy and its long-term stability. Consider the true state evolution $\bm{\omega}(t_{k+1}) = S(\bm{\omega}(t_k))$ and its model approximation $\hat{\bm{\omega}}(t_{k+1}) = \mathcal{F}(\hat{\bm{\omega}}(t_k))$. The prediction error $\bm{\epsilon}(t_k) = \hat{\bm{\omega}}(t_k) - \bm{\omega}^*(t_k)$ evolves approximately as:
\begin{equation} \label{eq:error_propagation_condensed}
\bm{\epsilon}(t_{k+1}) \approx \mathbf{A}_k \bm{\epsilon}(t_k) + \bm{e}_{\text{model}}(t_k),
\end{equation}
where $\mathbf{A}_k = \nabla S|_{\bm{\omega}^*(t_k)}$ is the system Jacobian and $\bm{e}_{\text{model}}(t_k) = \mathcal{F}(\bm{\omega}^*(t_k)) - S(\bm{\omega}^*(t_k))$ is the single-step model error. Let $\mathbf{P}_{\text{high}}$ be the projection operator onto a high-frequency subspace.

\begin{theorem}[\textit{High-Frequency Error Control and Prediction Stability}]
\label{thm:hf_error_control_condensed}
Assume the system Jacobian $\mathbf{A}_k$ amplifies high-frequency errors, with constants $C_A \geq 1$ and $\gamma_{\text{high}} > 1$, such that for $\bm{\epsilon}_{\text{high}}(t_k) = \mathbf{P}_{\text{high}}\bm{\epsilon}(t_k)$:
\begin{equation} \label{eq:system_hf_growth_condensed}
\|\mathbf{P}_{\text{high}} \mathbf{A}_k \bm{\epsilon}_{\text{high}}(t_k) \| \leq C_A \gamma_{\text{high}} \|\bm{\epsilon}_{\text{high}}(t_k) \|.
\end{equation}
If a model $\mathcal{F}$ ensures its single-step high-frequency model error is bounded by $\|\mathbf{P}_{\text{high}} \bm{e}_{\text{model}}(t_k) \| \leq \delta_{\text{HF}}$, then after $M$ autoregressive steps, the accumulated high-frequency error $\bm{\epsilon}_{\text{high}}(t_M)$ is bounded by:
\begin{equation} \label{eq:accumulated_hf_error_condensed}
\|\bm{\epsilon}_{\text{high}}(t_M)\| \leq (C_A \gamma_{\text{high}})^M \|\bm{\epsilon}_{\text{high}}(t_0)\| + \delta_{\text{HF}} \frac{(C_A \gamma_{\text{high}})^M - 1}{C_A \gamma_{\text{high}} - 1},
\end{equation}
for $C_A \gamma_{\text{high}} \neq 1$ and initial error $\bm{\epsilon}_{\text{high}}(t_0)$.
\end{theorem}

\noindent\textbf{Discussion of association with \ourmethod{}.} Theorem~\ref{thm:hf_error_control_condensed} implies that a smaller $\delta_{\text{HF}}$ is critical for mitigating error growth. The \ourmethod{} architecture, with its HDS mechanism, is designed to minimize $\delta_{\text{HF}}$ by explicitly addressing \textit{Spectral Bias}. This architectural focus on accurately capturing high-frequency dynamics leads to a significant reduction in the model-induced error term in Eq.~\eqref{eq:accumulated_hf_error_condensed}, thereby enhancing long-term stability.

\section{Proof of Theorem on High-Frequency Error Control}
\label{sec:appendix_proof}
This section provides a detailed proof for Theorem~\ref{thm:hf_error_control_condensed}.

\subsection{Preliminaries: Error Evolution Equation}
Let the true state evolution of a dynamical system be governed by $\bm{\omega}(t_{k+1}) = S(\bm{\omega}(t_k))$. A surrogate model approximates this as $\hat{\bm{\omega}}(t_{k+1}) = \mathcal{F}(\hat{\bm{\omega}}(t_k))$. The prediction error is $\bm{\epsilon}(t_k) = \hat{\bm{\omega}}(t_k) - \bm{\omega}^*(t_k)$. The error at the next step is:
\begin{align}
    \bm{\epsilon}(t_{k+1}) = \mathcal{F}(\hat{\bm{\omega}}(t_k)) - S(\bm{\omega}^*(t_k)).
\end{align}
A first-order Taylor expansion of $\mathcal{F}(\hat{\bm{\omega}}(t_k))$ around $\bm{\omega}^*(t_k)$ gives:
\begin{equation}
    \mathcal{F}(\hat{\bm{\omega}}(t_k)) \approx \mathcal{F}(\bm{\omega}^*(t_k)) + \nabla \mathcal{F}|_{\bm{\omega}^*(t_k)} \bm{\epsilon}(t_k).
\end{equation}
Let $\bm{e}_{\text{model}}(t_k) = \mathcal{F}(\bm{\omega}^*(t_k)) - S(\bm{\omega}^*(t_k))$ be the single-step model error. Assuming $\nabla \mathcal{F}|_{\bm{\omega}^*(t_k)} \approx \mathbf{A}_k = \nabla S|_{\bm{\omega}^*(t_k)}$, the error evolution is:
\begin{equation} \label{eq:appendix_error_propagation_condensed_proof}
    \bm{\epsilon}(t_{k+1}) \approx \mathbf{A}_k \bm{\epsilon}(t_k) + \bm{e}_{\text{model}}(t_k).
\end{equation}

\subsection{Detailed Proof}
\begin{proof}
Let $\bm{\epsilon}_{\text{high}}(t_k) = \mathbf{P}_{\text{high}}\bm{\epsilon}(t_k)$. Applying $\mathbf{P}_{\text{high}}$ to Eq.~\eqref{eq:appendix_error_propagation_condensed_proof}:
\begin{align}
    \bm{\epsilon}_{\text{high}}(t_{k+1}) \approx \mathbf{P}_{\text{high}}\mathbf{A}_k \bm{\epsilon}(t_k) + \mathbf{P}_{\text{high}}\bm{e}_{\text{model}}(t_k).
\end{align}
Taking the norm and applying the triangle inequality:
\begin{equation} \label{eq:appendix_hf_error_step1_proof}
    \|\bm{\epsilon}_{\text{high}}(t_{k+1})\| \leq \|\mathbf{P}_{\text{high}}\mathbf{A}_k \bm{\epsilon}(t_k)\| + \|\mathbf{P}_{\text{high}}\bm{e}_{\text{model}}(t_k)\|.
\end{equation}
We assume the amplification of high-frequency error is primarily driven by existing high-frequency components:
\begin{equation}
     \|\mathbf{P}_{\text{high}}\mathbf{A}_k \bm{\epsilon}(t_k)\| \leq C_A \gamma_{\text{high}} \|\bm{\epsilon}_{\text{high}}(t_k) \|.
\end{equation}
By the theorem's assumption, $\|\mathbf{P}_{\text{high}} \bm{e}_{\text{model}}(t_k) \| \leq \delta_{\text{HF}}$. Substituting these bounds into Eq.~\eqref{eq:appendix_hf_error_step1_proof}, we get the recursive inequality:
\begin{equation} \label{eq:appendix_recursive_inequality_proof}
    \|\bm{\epsilon}_{\text{high}}(t_{k+1})\| \leq C_A \gamma_{\text{high}} \|\bm{\epsilon}_{\text{high}}(t_k)\| + \delta_{\text{HF}}.
\end{equation}
Let $G = C_A \gamma_{\text{high}}$ and $x_k = \|\bm{\epsilon}_{\text{high}}(t_k)\|$. Unrolling the recursion $x_{k+1} \leq G x_k + \delta_{\text{HF}}$ for $M$ steps:
\begin{align*}
    x_M \leq G^M x_0 + \delta_{\text{HF}} \sum_{i=0}^{M-1} G^i.
\end{align*}
The geometric series sum is $\frac{G^M - 1}{G - 1}$ for $G \neq 1$. Substituting this back yields the bound:
\begin{equation}
    \|\bm{\epsilon}_{\text{high}}(t_M)\| \leq (C_A \gamma_{\text{high}})^M \|\bm{\epsilon}_{\text{high}}(t_0)\| + \delta_{\text{HF}} \frac{(C_A \gamma_{\text{high}})^M - 1}{C_A \gamma_{\text{high}} - 1}.
\end{equation}
This completes the proof.
\end{proof}

\subsection{Discussion and Association with \ourmethod{}}
Theorem~\ref{thm:hf_error_control_condensed} highlights two primary sources of high-frequency error accumulation: amplification of initial error and accumulation of single-step model error. Minimizing the single-step high-frequency error, $\delta_{\text{HF}}$, is critical. The \ourmethod{} architecture, with its HDS mechanism, is specifically designed to achieve a smaller $\delta_{\text{HF}}$ by mitigating spectral bias. This explains its superior performance in long-term, high-fidelity turbulence prediction.

\section{Numerical Experiments}

\begin{table*}[t]
\caption{Performance comparison on isotropic turbulence datasets measured by $L^2$ error at different prediction steps. Lower $L^2$ error values indicate better performance.}
    \label{tab:turbulence_results}
    \vskip 0.1in
    \centering
    \begin{footnotesize}
        \renewcommand{\multirowsetup}{\centering}
        \setlength{\tabcolsep}{3.2pt} 
        \newcolumntype{C}{>{\centering\arraybackslash}X} 
        \begin{tabularx}{\textwidth}{l@{\hspace{2pt}}l| *{5}{C} | *{3}{C}}
            \toprule
            \multicolumn{2}{l|}{\multirow{3}{*}{Model Category}} & \multicolumn{8}{c}{Datasets}  \\
            \cmidrule(lr){3-10}
            \multicolumn{2}{l|}{} & \multicolumn{5}{c}{Decaying Isotropic Turbulence} & \multicolumn{3}{c}{Forced Isotropic Turbulence} \\
            \cmidrule(lr){3-7} \cmidrule(lr){8-10}
            \multicolumn{2}{l|}{} & 1-step & 10-step & 40-step & 60-step & 99-step & 1-step & 10-step & 19-step \\
            \midrule
            \multicolumn{10}{l}{\textbf{Operator Learning Models}} \\
            \faPuzzlePiece & FNO \cite{li2021fourier} \ICLR{2021} & 0.0267 & 0.3550 & 1.7794 & 2.5634 & 3.1284 & 0.0118 & 0.1384 & 1.9832 \\
            \faPuzzlePiece & CNO \cite{raonic2023convolutional} \NeurIPS{2023} & 0.0407 & 1.7774 & 6.1085 & 7.7403 & 11.3015 & 0.0008 & 0.1227 & 1.5676 \\
            \faPuzzlePiece & LSM \cite{wu2023solving} \ICML{2023}  & 0.0046 & 0.3373 & 3.0698 & 4.2579 & 5.1127 & 0.0017 & 0.1287 & 2.0382 \\
            \faPuzzlePiece & NMO \cite{wu2024neural} \KDD{2024} & 0.0018 & 0.0602 & 1.7634 & 1.9832 & 2.1923 & 0.0002 & 0.0043 & 0.1873 \\
            \midrule
            \multicolumn{10}{l}{\textbf{Computer Vision Backbones}} \\
            \faCameraRetro & U-Net \cite{ronneberger2015u} \MICCAI{2015} & 0.0182 & 0.5810 & 2.8367 & 3.7935 & 4.6647 & 0.0007 & 0.0296 & 0.6583 \\
            \faCameraRetro & ResNet \cite{he2016deep} \CVPR{2016}  & 0.0098 & 0.4983 & 1.9823 & 2.9983 & 5.8743 & 0.0025 & 0.2424 &  2.3630\\
            \faCameraRetro & ViT \cite{dosovitskiy2021an} \ICLR{2021} & {0.0360} & {2.2588} & {5.4557} & {5.6154} & {5.6401} & 0.0074 & 0.2363 & {3.8732} \\
            \faCameraRetro & DiT \cite{peebles2023scalable} \ICCV{2023} & 0.0038 & 1.9832 & 2.7654 & 5.9862 & 9.9283 & 0.0007 & 0.0283 & 1.2731 \\
            \midrule
            \multicolumn{10}{l}{\textbf{Spatiotemporal Models}} \\
            \faFilm & ConvLSTM \cite{shi2015convolutional} \NIPS{2015} & 0.0374 & 0.5643 & 2.2374 & 3.9841 & 4.9381 & 0.0443 & 0.0687 & 1.2384 \\
            \faFilm & SimVP \cite{tan2022simvp} \CVPR{2022} & 0.0019 & 0.0621 & 2.1104 & 3.8504 & 5.0405 & 0.0002 & 0.0046 & 0.2231 \\
            \faFilm & PastNet \cite{wu2024pastnet} \MM{2024} & 0.0128 & 0.0415 & 0.5689 & 1.7574 & 2.3837 &  0.0073 & 0.0348 & 0.6173 \\
            \midrule
            \rowcolor{gray!10} 
            \faTrophy & ~\textbf{\ourmethod{}} & \textbf{{0.0002}} & \textbf{{0.0048}} & \textbf{{0.0807}} & \textbf{{0.2157}} & \textbf{{0.6171}} & \textbf{{0.0001}} & \textbf{{0.0019}} & \textbf{{0.1257}} \\
            \rowcolor{gray!10} 
            & Promotion & 88.89\% & 92.29\% & 85.82\% & 89.13\% & 80.30\% & 50.00\% & 58.70\% & 43.66\% \\
            \bottomrule
        \end{tabularx}
    \end{footnotesize}
\end{table*}

\begin{figure*}[t]
    \centering
    \includegraphics[width=\textwidth]{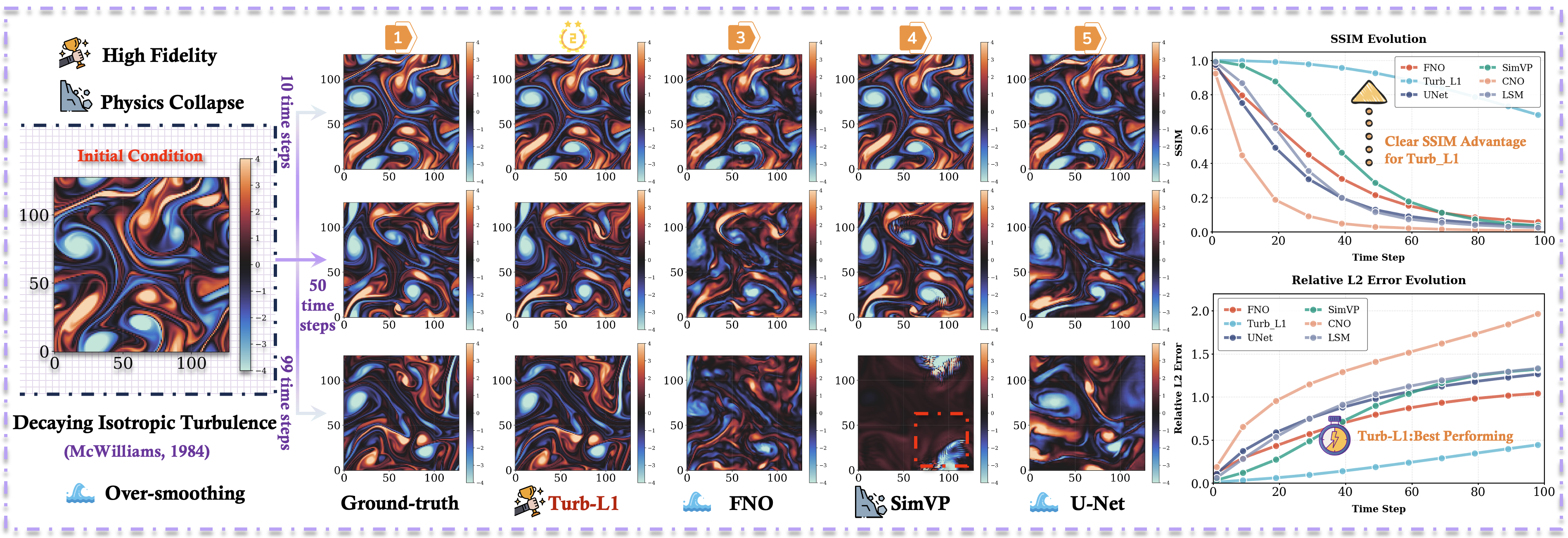} 
    \caption{\textbf{Long-term prediction on 2D Decaying Isotropic Turbulence.}
        \textit{(Left)} Initial condition.
        \textit{(Center Columns 1-5)} Vorticity fields at $t=10, 50, 99$ for Ground-truth, ~\ourmethod{}, FNO, SimVP, and U-Net.  ~\ourmethod{} maintains high fidelity, while FNO/U-Net show over-smoothing and SimVP (collapse icon) exhibits instability, underscoring  ~\ourmethod{}'s resilience to spectral bias.
        \textit{(Right)} Quantitative evaluation: {(Top Right.)} SSIM Evolution, where  ~\ourmethod{} excels in structural similarity. {(Bottom Right.)} Relative $L^2$ Error Evolution, showing  ~\ourmethod{}'s superior accuracy and stability.}
\label{fig:Results_figure1}
\end{figure*}

\subsection{Experiment Setup}
\noindent\textbf{Datasets}~\quad We test \ourmethod{} on two classic datasets of two-dimensional isotropic turbulence, a canonical problem in fluid dynamics. These datasets are designed to evaluate the model's capability to capture complex chaotic dynamics under different physical regimes. \ding{64}~\textbf{Dataset I: Forced Isotropic Turbulence Benchmark.}
This scenario follows the Navier-Stokes equation benchmark proposed in Li et al.~\cite{li2021fourier}. The kinematic viscosity is set to $\nu = 10^{-5}$. The initial vorticity field, $\omega_0$, is sampled from a zero-mean Gaussian random field with a covariance operator $(-\Delta + \tau^2 I)^{-\alpha/2}$, where $\Delta$ is the Laplacian operator, $I$ is the identity operator, and $\tau$ and $\alpha$ are parameters controlling the spectral shape. The energy density $E(k)$ exhibits a decay law of $E(k) \sim (k^2 + \tau^2)^{-\alpha}$. The flow is driven by a fixed, low-wavenumber external force, and no additional drag effects are considered. \ding{64}~\textbf{Dataset II: Decaying Isotropic Turbulence (McWilliams, 1984).}~\cite{mcwilliams1984emergence}
This scenario simulates the phenomenon of decaying turbulence. The power spectrum of the initial streamfunction, characterized by $|\hat{\psi}(\mathbf{k})|^2 \sim k^{-1}(\tau_0^2 + (k/k_0)^4)^{-1}$, where $k_0$ is the characteristic wavenumber and $\tau_0$ is a spectral shape parameter, defines the initial flow field. The initial conditions facilitate a slow energy decay, and the enstrophy density evolves to exhibit features akin to a Kolmogorov energy cascade. The turbulence is unforced and evolves naturally. 

\noindent\textbf{Backbones}~\quad To comprehensively evaluate the performance of our proposed method, we select three categories of representative deep learning models as baselines for comparison. \faPuzzlePiece \; \textbf{Operator Learning Models}. These models aim to directly learn mappings between infinite-dimensional function spaces. We choose several mainstream and cutting-edge methods from this field, including: FNO~\cite{li2021fourier}, CNO~\cite{raonic2023convolutional}, LSM~\cite{wu2023solving}, and NMO~\cite{wu2024neural}. \faCameraRetro \; \textbf{Classic Computer Vision Backbones}. These models demonstrate strong feature extraction capabilities in image processing and computer vision tasks. Specific models include: U-Net~\cite{ronneberger2015u}, ResNet~\cite{he2016deep}, ViT~\cite{dosovitskiy2021an}, and DiT~\cite{peebles2023scalable}. \faFilm \; \textbf{Spatiotemporal Prediction Models}. These models are specifically designed to process and predict sequential data with temporal and spatial dependencies. Our selected representative models are: ConvLSTM~\cite{shi2015convolutional}, SimVP~\cite{tan2022simvp}, and PastNet~\cite{wu2024pastnet}. 

\noindent\textbf{Experiment settings}~\quad We train all models in this paper on a server equipped with eight NVIDIA A100 GPUs, each with 40GB of memory. We use the DistributedDataParallel mode from the PyTorch framework for distributed training to accelerate convergence. Model inference is performed on a single NVIDIA A100 GPU. The software environment for our experiments is based on Python 3.8. We use PyTorch version 1.8.1 (CUDA 11.1), TorchVision version 0.9.1. During the training process, we consistently use the following hyperparameter configuration: the batch size is set to 20, the total number of epochs is 500, and the initial learning rate is 0.001. To ensure the reproducibility of our experiments, we fix all random seeds to 42. 

\noindent\textbf{Metrics}~\quad To comprehensively evaluate our proposed method, we employ five metrics. As follows:

\noindent \ding{224} \textit{\textbf{$L^2$ Error~\cite{heydari2012error}.}}
Measures the $L^2$ norm of the error field, which is the difference between the predicted field ($\hat{\omega}(\cdot, t)$) and the ground truth field ($\omega(\cdot, t)$).
\begin{equation}
\begin{aligned}
    & L^2 \text{ Error}(t) = \|\hat{\omega}(\cdot, t) - \omega(\cdot, t)\|_{L^2} \\ & =
     \sqrt{\sum_{i=1}^{N_x} \sum_{j=1}^{N_y} (\hat{\omega}_{i,j}(t) - \omega_{i,j}(t))^2}
     \end{aligned}
\end{equation}

\noindent \ding{224} \textit{\textbf{Structural Similarity Index Measure (SSIM)~\cite{sampat2009complex}.}}
Assesses the perceptual similarity between two fields (prediction $u$, ground truth $v$) considering local means ($\mu$), variances ($\sigma^2$), and covariance ($\sigma_{uv}$).
\begin{equation}
    \text{SSIM}(u, v) = \frac{(2\mu_u\mu_v + c_1)(2\sigma_{uv} + c_2)}{(\mu_u^2 + \mu_v^2 + c_1)(\sigma_u^2 + \sigma_v^2 + c_2)}
\end{equation}
where $c_1, c_2$ are stabilization constants. The overall SSIM is the mean of local SSIM values.

\noindent \ding{224} \textit{\textbf{Relative $L^2$ Error~\cite{chen2010least}.}}
Normalizes the $L^2$ norm of the error field by the $L^2$ norm of the ground truth field.
\begin{equation}
\begin{aligned}
    & \text{Relative } L^2 \text{ Error}(t) = \frac{\|\hat{\omega}(\cdot, t) - \omega(\cdot, t)\|_{L^2}}{\|\omega(\cdot, t)\|_{L^2}} \\& = \frac{\sqrt{\sum_{i=1}^{N_x} \sum_{j=1}^{N_y} (\hat{\omega}_{i,j}(t) - \omega_{i,j}(t))^2}}{\sqrt{\sum_{i=1}^{N_x} \sum_{j=1}^{N_y} (\omega_{i,j}(t))^2}} 
\end{aligned}
\end{equation}

\noindent \ding{224} \textit{\textbf{Enstrophy Spectrum (Density), $E_Z(k,t)$~\cite{verkley2009energy}.}}
Describes the distribution of enstrophy (mean squared vorticity) across different wavenumbers $k$. Total enstrophy $Z(t) = \int_0^\infty E_Z(k,t) dk$. For 2D fields, it's computed from the Fourier transform of vorticity $\hat{\omega}(\mathbf{k}',t)$ over annular shells in Fourier space, where $A$ is the domain area:
\begin{equation}
    E_Z(k,t) \approx \frac{1}{\Delta k} \sum_{k \le |\mathbf{k}'| < k+\Delta k} \frac{1}{2A} |\hat{\omega}(\mathbf{k}', t)|^2 \cdot (2\pi k')
\end{equation}

\noindent \ding{224} \textit{\textbf{Normalized Spectral Error~\cite{bazar2016evaluating}.}}
Quantifies the relative difference between the predicted enstrophy spectrum ($E_{Z,\text{pred}}(k,t)$) and the ground truth enstrophy spectrum ($E_{Z,\text{GT}}(k,t)$) at each wavenumber $k$.
\begin{equation}
    \text{Normalized Spectral Error}(k,t) = \frac{E_{Z,\text{pred}}(k,t) - E_{Z,\text{GT}}(k,t)}{E_{Z,\text{GT}}(k,t)}
\end{equation}

\begin{figure*}[t]
    \centering
    \includegraphics[width=\textwidth]{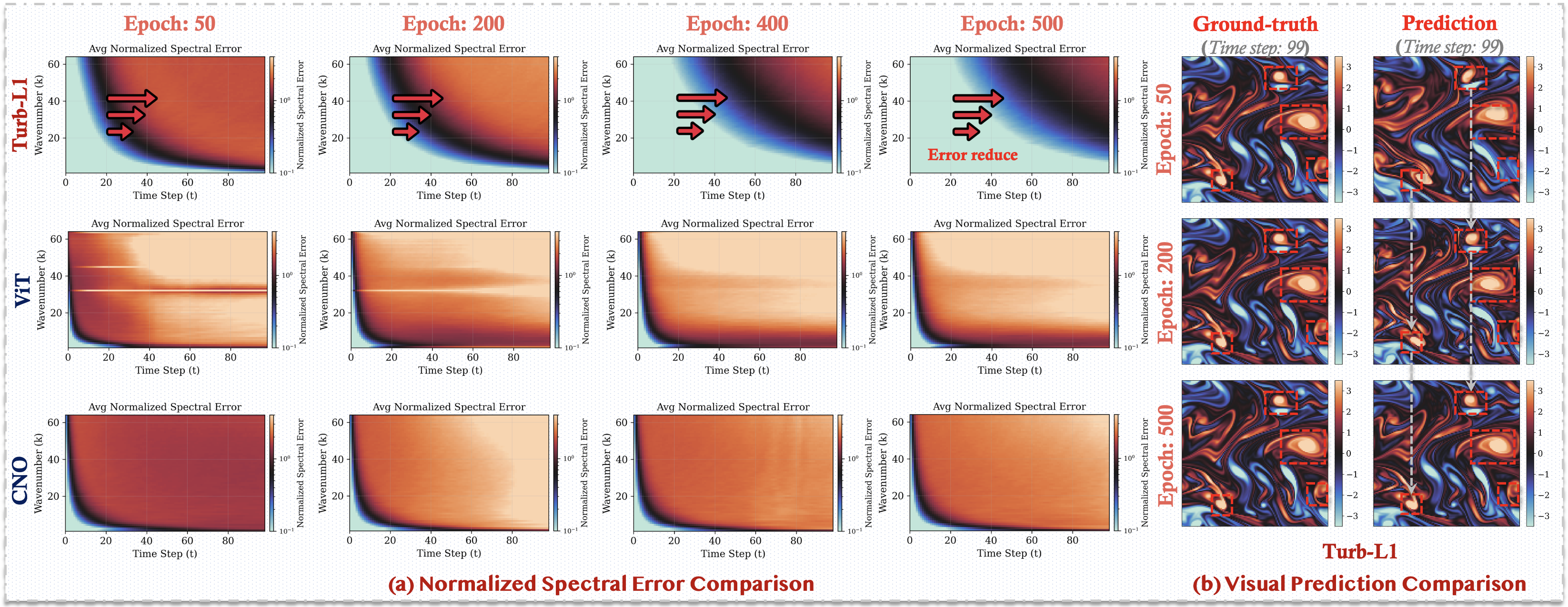}
\caption{Evolution of spectral error and visual prediction improvement for \ourmethod{} during training. (a) Normalized spectral error for \ourmethod{}, ViT, and CNO at different training epochs. \ourmethod{} shows significantly reduced error in high-wavenumber regions with training. (b) Visual comparison of \ourmethod{}'s long-term predictions at early (Epoch: 50) and late (Epoch: 500) training stages, demonstrating enhanced capturing of high-frequency details and vortex structures.}
\label{fig:epoch_change}
\end{figure*}

\subsection{Main results}
Table \ref{tab:turbulence_results} presents the performance achieved across different baselines over various prediction steps. We summarize the key takeaways as follows:

\noindent\textbf{Takeaway \ding{202}: \ourmethod consistently achieves state-of-the-art performance across diverse benchmarks and prediction steps.}~\quad Across both Dataset I and II, \ourmethod achieves significantly lower $L^2$ errors compared to all three mainstream method categories (\faPuzzlePiece, \faCameraRetro \; and \faFilm). Notably, on Dataset I, \ourmethod demonstrates exceptional performance with an error of only 0.6171, substantially outperforming the second-best baseline NMO (2.1923). This superior performance remains consistently observable in Dataset II (0.1257 vs. 0.1873). More generally, \ourmethod attains overall error reductions ranging from $43.66\%\sim92.29\%$ across different prediction steps, conclusively validating our \ourmethod effectiveness for long-term forecasting.

\noindent\textbf{Takeaway \ding{203}: \ourmethod{} demonstrates excellent visual fidelity and physical consistency in long-term predictions. It effectively overcomes over-smoothing and simulation collapse, issues that spectral bias causes.} As Figure~\ref{fig:Results_figure1} illustrates,  \ourmethod{}'s predictions closely match the ground truth even after 99 time steps, clearly reproducing complex vortex structures and fine flow features. This performance stems from its core Hierarchical Dynamics Synthesis mechanism, which precisely captures full-spectrum dynamic information, particularly high-frequency details. In contrast, FNO exhibits significant over-smoothing due to spectral bias and loses critical high-frequency information. SimVP, on the other hand, experiences physics collapse from accumulated high-frequency error amplification, producing unrealistic artifacts.  \ourmethod{}'s SSIM and relative L2 error curves further confirm its significant advantages in maintaining structural similarity and long-term prediction accuracy. This fully demonstrates the method's robustness and high fidelity in tracking complex multi-scale turbulence dynamics.

\subsection{Training Dynamics: How \ourmethod{} Progressively Overcomes Spectral Bias}
\textbf{Takeaway \ding{204}: Figure~\ref{fig:epoch_change} shows \ourmethod{}'s progress in overcoming spectral bias and improving high-frequency prediction during training.}
The left plots display normalized spectral error: \ourmethod{}'s high-wavenumber error significantly decreases as epochs increase from 50 to 500, indicating its $\mathcal{HDS}$ mechanism effectively learns high-frequency details.
In contrast, ViT and CNO consistently show large high-frequency errors, highlighting their limitations in overcoming spectral bias.
The visual comparisons on the right confirm this: \ourmethod{} trained to Epoch 500 accurately reproduces fine vortex structures and curled edges, which are blurred at Epoch 50.
This improved vortical detail capture reflects its enhanced high-frequency prediction.
Thus, \ourmethod{}'s architecture-guided training effectively mitigates spectral bias, enabling high-fidelity tracking of crucial vortex dynamics.

\subsection{\ourmethod{}'s Superiority in Enstrophy Spectrum Prediction}
\begin{figure}
 \centering
 \includegraphics[width=0.48\textwidth]{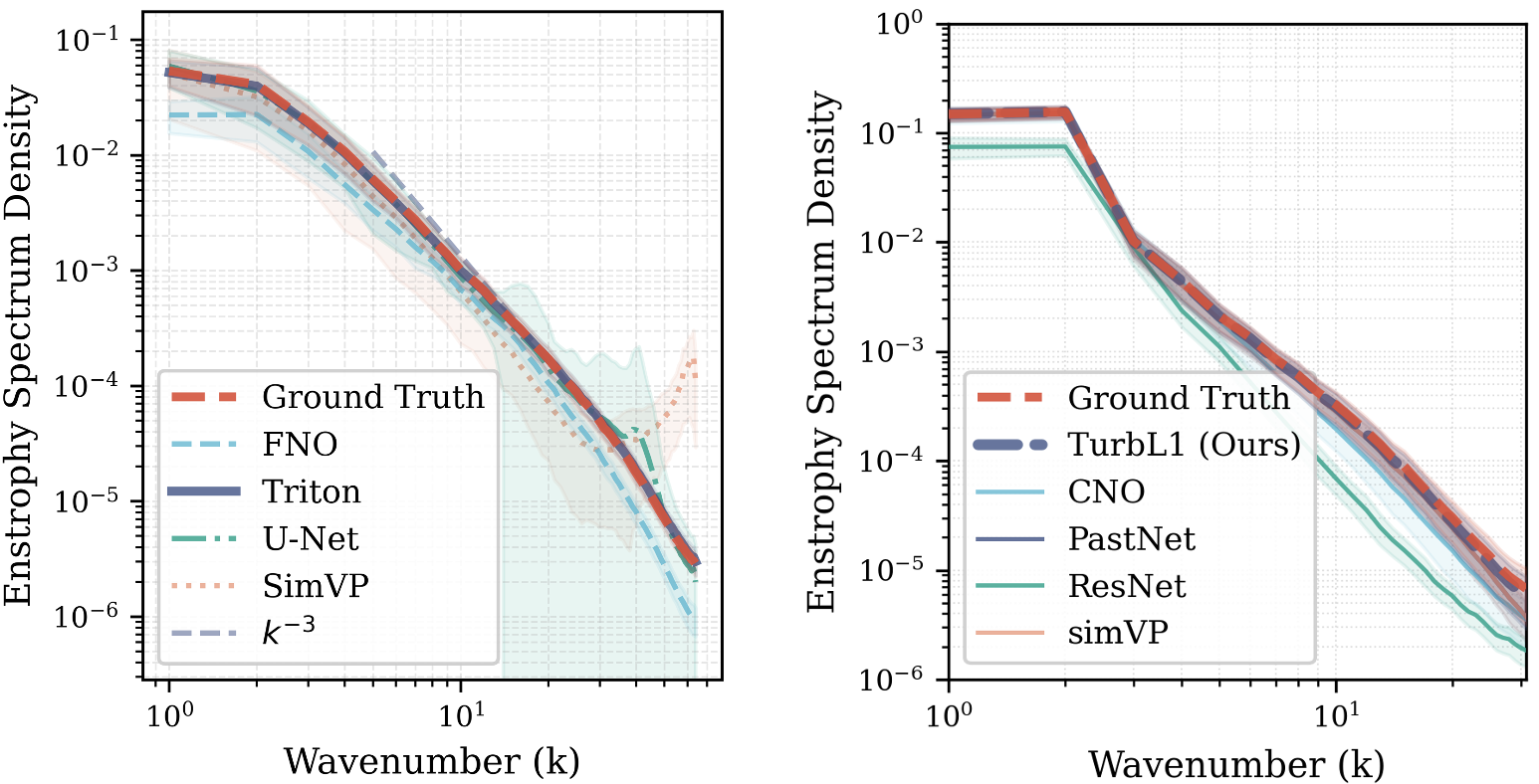}
\caption{Comparison of enstrophy spectrum analysis. Left plot is Dataset I, Right plot is Dataset II.}
\label{fig:enhanced_enstrophy_spectrum}
\end{figure}

\begin{figure}[t]
    \centering
    \includegraphics[width=0.48\textwidth]{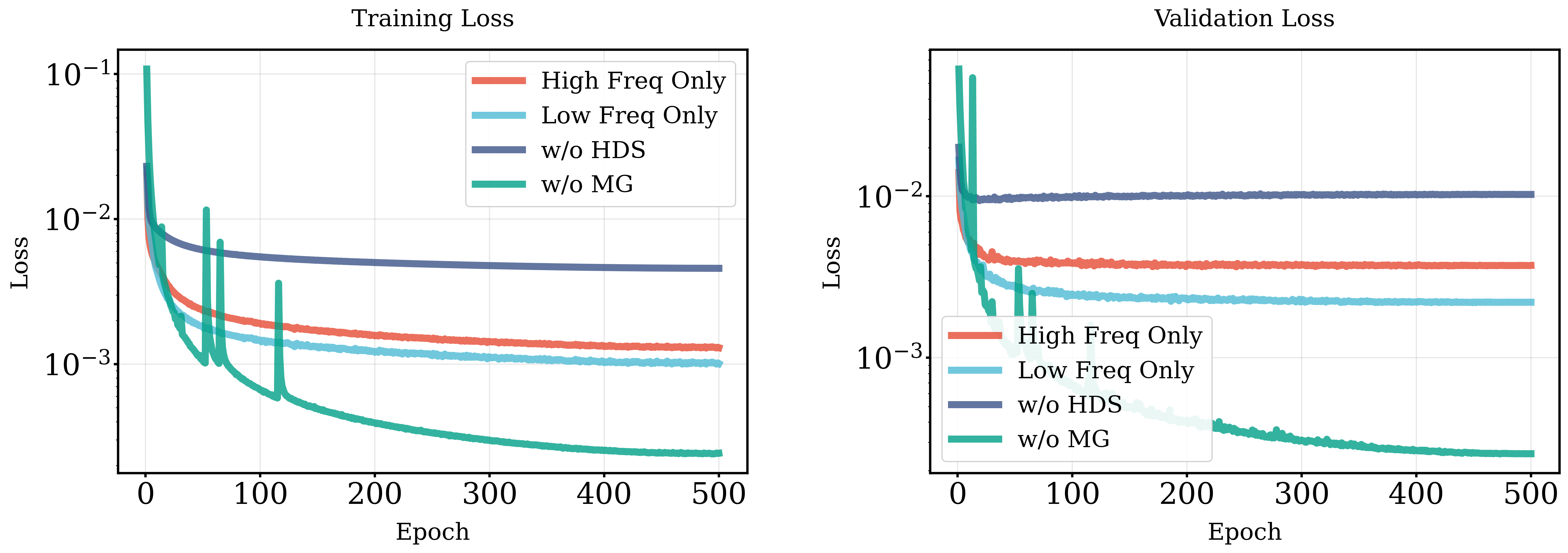} 
   \caption{Training and validation loss curves for different ablation variants of \ourmethod{} on the Decaying Isotropic Turbulence dataset over 500 epochs.
(\textbf{Left}) Training loss. (\textbf{Right}) Validation loss.
The legend indicates: `High Freq Only' (model primarily processing high-frequency components), `Low Freq Only' (model primarily processing low-frequency components), `w/o HDS' (model without the Hierarchical Dynamics Synthesis mechanism), and `w/o MG' (model without the Multi-Grid architecture but retaining HDS).
The `w/o HDS' variant exhibits the highest loss, highlighting the critical role of the HDS mechanism. The `w/o MG' variant, which retains HDS, achieves the lowest loss among the depicted ablated models. The y-axis is on a logarithmic scale.}
\label{fig:ablation_loss_curves}
\end{figure}

\begin{figure*}[h]
    \centering
    \includegraphics[width=\textwidth]{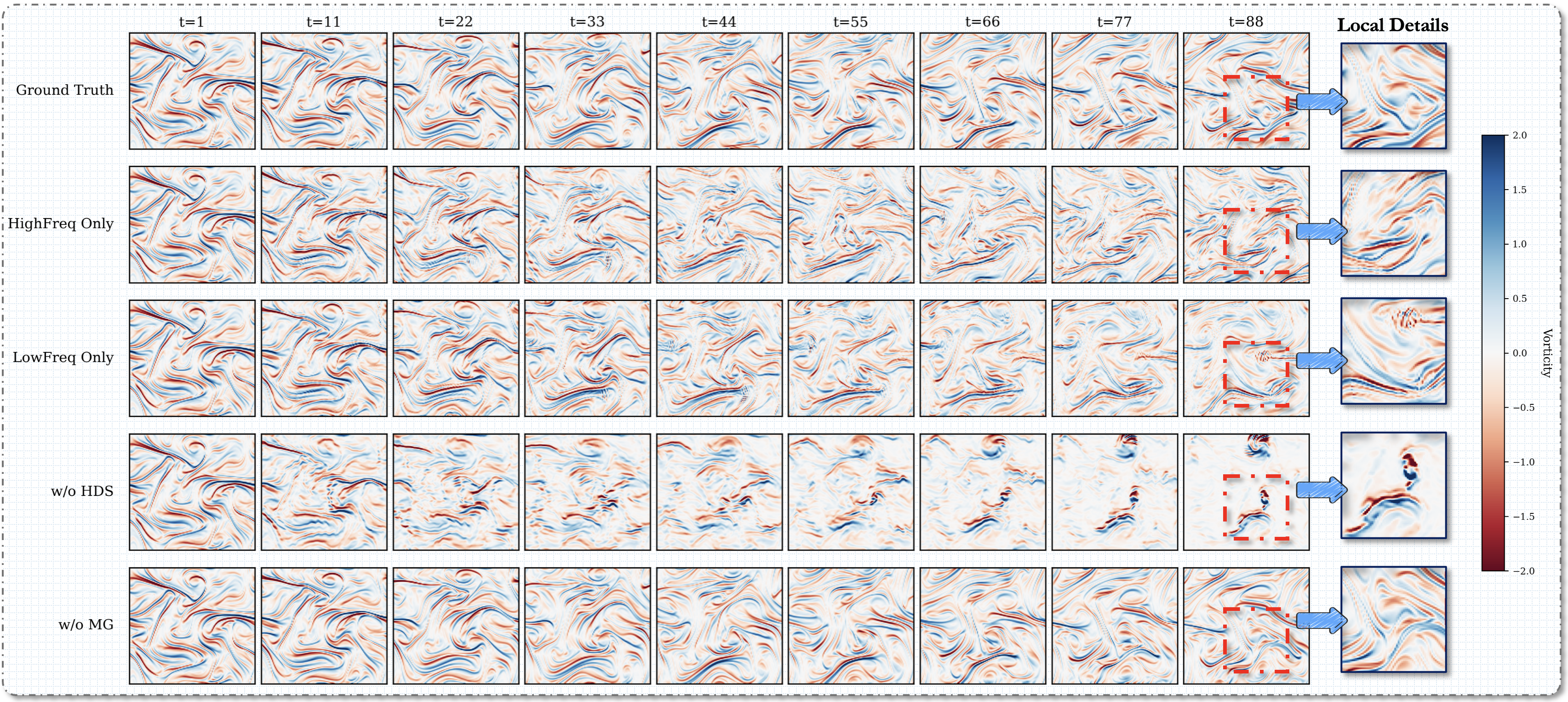} 
\caption{Visual comparison of predicted vorticity fields ($\omega$) by different ablation variants of \ourmethod{} against the Ground Truth for Decaying Isotropic Turbulence.
Snapshots are shown at nine time instances from $t=1$ to $t=88$.
Rows display results from: Ground Truth, `HighFreq Only', `LowFreq Only', `w/o HDS' (without Hierarchical Dynamics Synthesis), and `w/o MG' (with HDS but without Multi-Grid architecture).
The `w/o MG' variant qualitatively best reproduces the evolution of turbulent structures and maintains high fidelity to the Ground Truth over extended periods, significantly outperforming other ablated models. The `w/o HDS' variant fails to capture any physically realistic dynamics, while `HighFreq Only' and `LowFreq Only' variants exhibit excessive smoothing and loss of fine-scale details over time.}
\label{fig:ablation_vorticity_visualization}
\end{figure*}

\begin{figure}[h]
    \centering
    \includegraphics[width=0.45\textwidth]{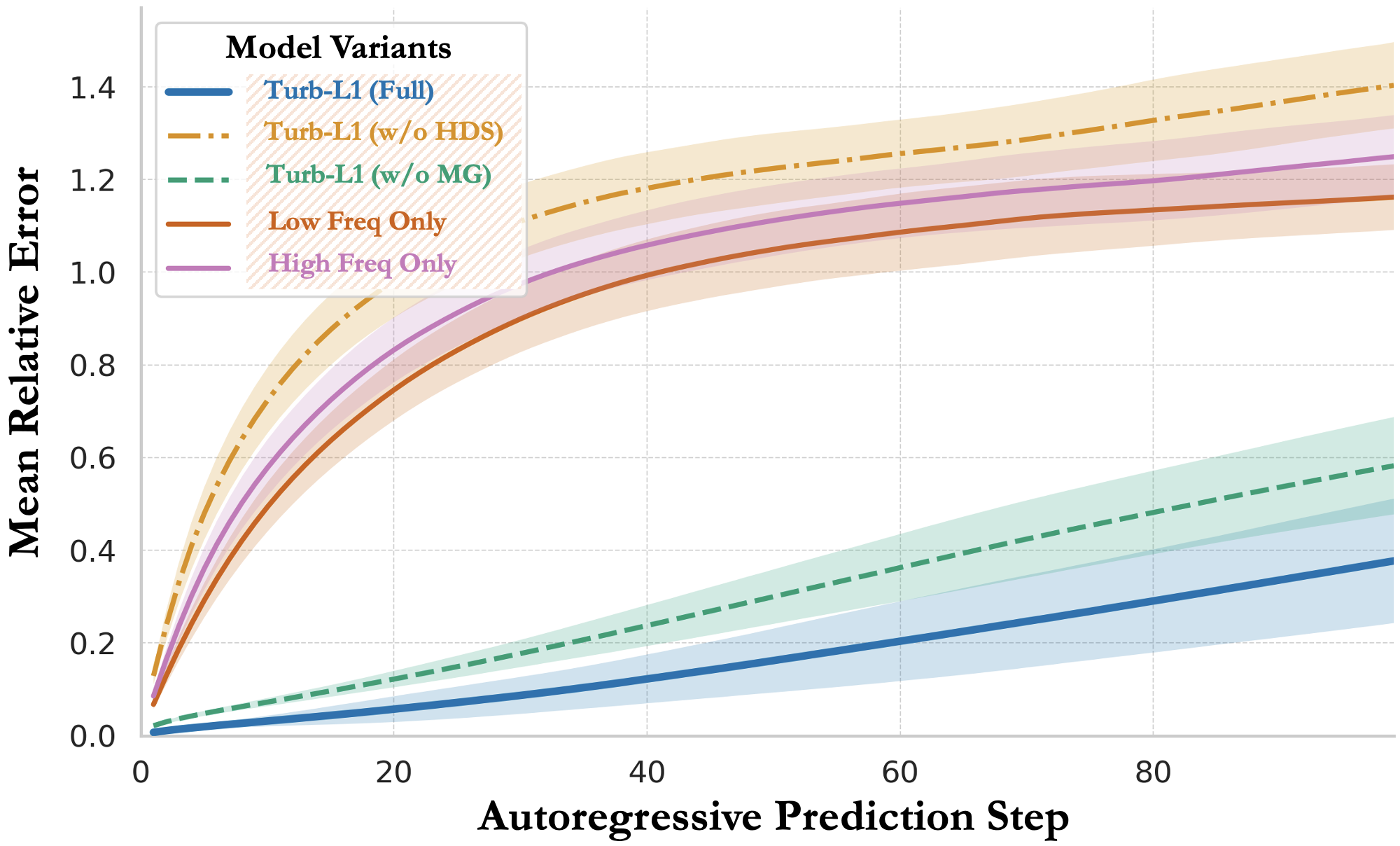} 
\caption{Ablation study on the long-term predictive performance of \ourmethod{} and its variants. The plot shows the Mean Relative Error accumulated over 100 autoregressive prediction steps.
    The full \ourmethod{} model (\textbf{solid blue line}) significantly outperforms all ablated versions, maintaining a very low and stable error growth rate.
    Removing the Multi-Grid architecture (\textbf{dashed green line, w/o MG}) leads to a noticeable increase in error, though it remains more stable than other variants.
    Crucially, removing the core HDS mechanism (\textbf{dot-dashed orange line, w/o HDS}) or relying on only a single frequency band (\textbf{Low Freq Only} or \textbf{High Freq Only}) results in catastrophic performance degradation, with errors rapidly accumulating.
    These results demonstrate that the synergy between the multi-grid architecture and the Hierarchical Dynamics Synthesis (HDS) mechanism is essential for achieving the long-term stability and accuracy of \ourmethod{}.}
\label{fig:ablation_longterm_performance}
\end{figure}

\begin{table}[t] 
    \caption{Ablation study on the McWilliams 2D decaying turbulence dataset. We evaluate the performance of \textbf{\ourmethod{}} against its variants by removing key components: the Hierarchical Dynamics Synthesis (HDS) module and the Multi-Grid (MG) architecture. The variants `w/o High-Freq.` and `w/o Low-Freq.` represent models trained without the respective frequency-processing paths. Lower Relative L2 Error and higher SSIM indicate better performance. The results clearly demonstrate that every component is crucial for achieving state-of-the-art performance.}
    \label{tab:ablation_study}
    \vskip 0.1in
    \centering
    \begin{footnotesize}
        \renewcommand{\multirowsetup}{\centering}
        \setlength{\tabcolsep}{4.5pt} 
        
        \newcolumntype{C}[1]{>{\centering\arraybackslash}p{#1}}

        \begin{tabular}{l | C{1.1cm} C{1.1cm} | C{1.1cm} C{1.1cm}}
            \toprule
            \multirow{3}{*}{Model} & \multicolumn{4}{c}{Prediction Steps} \\
            \cmidrule(lr){2-5}
            & \multicolumn{2}{c}{Step 50} & \multicolumn{2}{c}{Step 99} \\
            \cmidrule(lr){2-3} \cmidrule(lr){4-5}
            & R. $L^2 (\downarrow)$ & SSIM $(\uparrow)$ & R. $L^2 (\downarrow)$ & SSIM $(\uparrow)$ \\
            \midrule
            
            \ourmethod{} (w/o HDS)     & 1.2240 & 0.0451 & 1.4032 & 0.0378 \\
            \ourmethod{} (w/o MG)      & 0.3001 & 0.8475 & 0.5825 & 0.5596 \\
            w/o High-Freq.             & 1.0494 & 0.1098 & 1.1620 & 0.0519 \\
            w/o Low-Freq.              & 1.1116 & 0.0822 & 1.2493 & 0.0396 \\
            
            \midrule
            \rowcolor{gray!15} 
            \textbf{\ourmethod{} (Full)} & \textbf{0.1622} & \textbf{0.9496} & \textbf{0.3772} & \textbf{0.7797} \\
            \bottomrule
        \end{tabular}
    \end{footnotesize}
\end{table}

\begin{figure*}[t]
    \centering
    \includegraphics[width=0.9\textwidth]{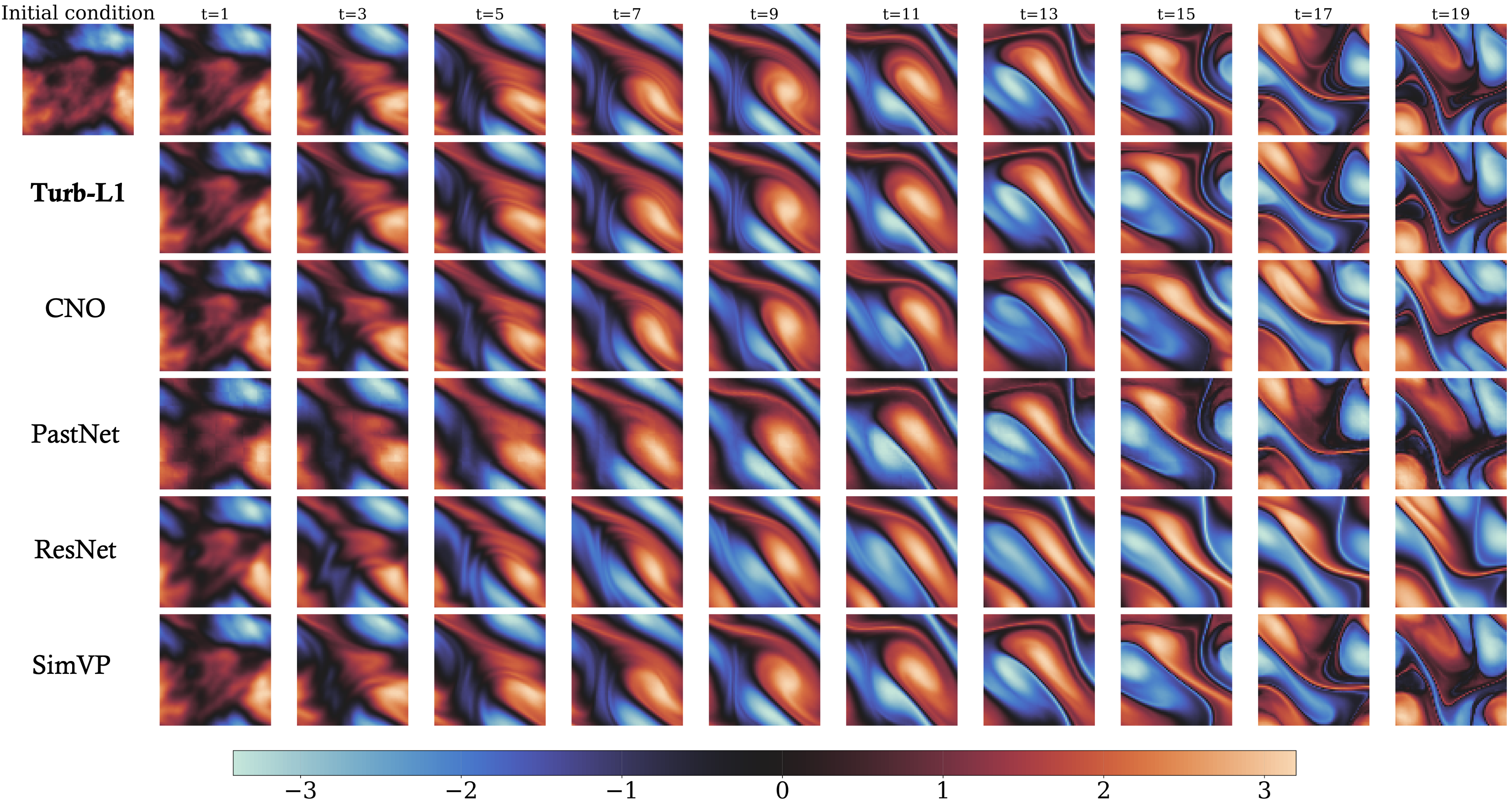}
\caption{Qualitative comparison of vorticity field ($\omega$) predictions for Forced Isotropic Turbulence by \ourmethod{} (Turb-L1) and baseline methods.
The first row displays the initial condition and subsequent snapshots representing the ground truth evolution.
Predictions are shown for time steps $t=1, 3, \dots, 19$.
\ourmethod{} maintains high fidelity to the ground truth evolution, accurately capturing fine-scale structures and complex interactions throughout the sequence.
In contrast, baseline methods such as CNO, PastNet, ResNet, and SimVP exhibit varying degrees of excessive smoothing, loss of high-frequency details, and structural distortions, particularly at later time steps.
This visualization further underscores the capability of \ourmethod{} to overcome spectral bias and achieve robust long-term turbulence forecasting.}
\label{fig:forced_turbulence_visualization} 
\end{figure*}

\textbf{Takeaway \ding{205}:} The enstrophy spectrum analysis further reveals \ourmethod{}'s significant effectiveness in overcoming spectral bias, as illustrated in Figure~\ref{fig:enhanced_enstrophy_spectrum} Compared to several baseline models, the enstrophy spectrum predicted by \ourmethod{} exhibits high consistency with the Ground Truth across the entire range, from low to high wavenumbers.
Many comparative methods, such as FNO and CNO, show a significant underestimation of energy or excessive smoothing in the high-wavenumber regions, with their spectral lines falling far below the true values. This is a typical manifestation of spectral bias, where models neglect high-frequency details.
In more extreme cases like SimVP, an anomalous energy accumulation even occurs in the high-frequency region, indicating the model's inability to stably handle high-frequency dynamics.
In contrast, \ourmethod{} accurately captures the crucial $k^{-3}$ energy cascade characteristic of turbulence. Its precise agreement, especially in the high-frequency part, demonstrates that it effectively learns and preserves small-scale vortex information vital for turbulence evolution, facilitated by mechanisms like Hierarchical Dynamics Synthesis.
This full-spectrum, high-fidelity prediction capability is a key factor in \ourmethod{}'s ability to achieve long-term, physically consistent turbulence simulations and confirms its success in mitigating spectral bias from a frequency-domain perspective.

\subsection{Ablation Study}

\subsubsection{Ablation Study Setup for Decaying Isotropic Turbulence}
\label{sec:ablation_setup_decaying}

To rigorously evaluate the contributions of the core components within our proposed \ourmethod{} architecture and to understand its mechanisms for overcoming spectral bias, we conducted a series of ablation studies on the Decaying Isotropic Turbulence benchmark dataset. This dataset, characterized by its unforced evolution and the natural emergence of a Kolmogorov-like energy cascade, provides a challenging scenario for testing long-term prediction fidelity, particularly concerning the preservation of fine-scale vortex structures and high-frequency dynamics.

Our ablation analysis focuses on dissecting the impact of the two primary architectural innovations in \ourmethod{}: the \texttt{Hierarchical Dynamics Synthesis (HDS)} mechanism and the Multi-Grid (MG) framework. Additionally, we explore the model's ability to leverage information across the frequency spectrum by creating variants that are restricted to primarily processing either high-frequency or low-frequency components.

The following ablated versions of \ourmethod{} were designed and evaluated:

\begin{enumerate}
    \item \textbf{\ourmethod{} w/o HDS (Without Hierarchical Dynamics Synthesis):}
    In this variant, the core HDS module is entirely removed from the \ourmethod{} architecture. The multi-grid encoder and decoder structures are retained, and a baseline dynamics evolution module (e.g., a stack of three standard convolutional layers) replaces the HDS mechanism. This setup aims to quantify the direct impact of the explicit multi-frequency synthesis performed by HDS.

    \item \textbf{\ourmethod{} w/o MG (Without Multi-Grid Architecture):}
    This variant removes the multi-grid projection and prolongation operators from both the encoder and decoder. The model operates on a single, fixed-resolution latent space. The HDS mechanism, however, is retained and adapted to function within this single-grid framework. This configuration is designed to isolate the benefits derived from the hierarchical, multi-scale processing enabled by the MG architecture.

    \item \textbf{\ourmethod{} (High-Freq Only Path in HDS):}
    To assess the model's reliance on high-frequency information, we modified the HDS module within the full \ourmethod{} (including MG). Specifically, the pathways responsible for processing low and middle-frequency dynamic primitives within HDS (i.e., $\mathcal{T}_{\omega_L}$ and $\mathcal{T}_{\omega_M}$ as described in Section~\ref{sec:hds_corrected}) are disabled. The model is thus forced to primarily rely on the high-frequency synthesis path (i.e., $\mathcal{T}_{\omega_H}$) for evolving the turbulent state.

    \item \textbf{\ourmethod{} (Low-Freq Only Path in HDS):}
    Complementary to the "High-Freq Only" variant, this setup modifies the HDS module to predominantly utilize the low-frequency synthesis pathway (i.e., $\mathcal{T}_{\omega_L}$). The high and middle-frequency pathways are disabled, compelling the model to base its predictions primarily on the evolution of large-scale, low-frequency structures.
\end{enumerate}

For all ablation variants, the training procedure, including hyperparameters, loss functions, and the number of training epochs, was kept consistent with the training of the full \ourmethod{} model to ensure a fair comparison.

\subsubsection{Analysis of Ablation Study Results}
\label{sec:analysis_ablation_results}

The effectiveness of the core components within \ourmethod{}, namely the Hierarchical Dynamics Synthesis (HDS) mechanism and the Multi-Grid (MG) architecture, was further investigated through a comprehensive ablation study. The results, presented in Figure~\ref{fig:ablation_loss_curves} (training and validation losses) and Figure~\ref{fig:ablation_longterm_performance} (long-term MSE and SSIM), provide clear insights into the contribution of each component.

\noindent\textbf{Critical Role of Hierarchical Dynamics Synthesis (HDS).}
The HDS mechanism stands out as the most critical component for the model's performance. As shown in Figure~\ref{fig:ablation_loss_curves}, the variant `w/o HDS` (model without HDS, dark blue curve in loss plots) exhibited by far the highest training and validation losses, saturating at approximately $5 \times 10^{-3} - 10^{-2}$. This poor learning capability translated directly into dismal long-term prediction accuracy (Figure~\ref{fig:ablation_longterm_performance}, teal curve for `w/o HDS`): its MSE rapidly escalated beyond 5.0, and its SSIM plummeted to near zero within a few time steps. This starkly contrasts with variants incorporating HDS. This underscores that the explicit, multi-path synthesis of dynamics across different frequency scales, as performed by HDS, is indispensable for learning the complex, multi-scale nature of turbulence and for overcoming the spectral bias that plagues conventional architectures.

\noindent\textbf{Contribution of the Multi-Grid (MG) Architecture.}
The `w/o MG` variant (model with HDS but without the MG architecture, teal curve in Figure~\ref{fig:ablation_loss_curves}, light purple curve in Figure~\ref{fig:ablation_longterm_performance}) demonstrated the best performance among all ablated models shown in the figures. Its training and validation losses were the lowest (Figure~\ref{fig:ablation_loss_curves}), and it achieved significantly superior long-term prediction stability, maintaining an MSE around 0.5-0.6 and an SSIM around 0.7 even at $t=100$ (Figure~\ref{fig:ablation_longterm_performance}). While the full \ourmethod{} model (which includes both HDS and MG, and whose performance surpasses this `w/o MG` variant) is not depicted in these specific ablation comparison figures, the strong performance of the `w/o MG` variant itself highlights the robustness conferred by the HDS mechanism. The further improvement achieved by the full \ourmethod{} indicates that the MG architecture provides an additional synergistic benefit by facilitating more effective hierarchical feature extraction and cross-scale interaction modeling, thereby enhancing the overall predictive capabilities of the HDS module.

\noindent\textbf{Necessity of Full-Spectrum Information Processing.}
The variants `High Freq Only` and `Low Freq Only` (processing predominantly high or low-frequency components, respectively) both showed substantially poorer performance than the `w/o MG` variant (which processes the full spectrum via HDS). In Figure~\ref{fig:ablation_loss_curves}, their losses saturated at higher levels (orange-red and light blue curves). More critically, their long-term prediction fidelity was severely compromised (Figure~\ref{fig:ablation_longterm_performance}, pink and olive green curves), with high MSE values and SSIM scores rapidly decaying to zero. The `Low Freq Only` variant performed marginally better than the `High Freq Only` variant in terms of loss and long-term metrics, suggesting that large-scale structures might be somewhat easier to capture or contribute more to these aggregate error metrics in decaying turbulence. However, neither approach was sufficient. This confirms that accurately predicting turbulence evolution necessitates a model capable of capturing and synthesizing information across the entire frequency spectrum, a capability central to the design of our HDS mechanism.

In summary, these ablation studies validate our architectural choices. The HDS mechanism is paramount for achieving high-fidelity turbulence prediction by explicitly addressing spectral bias. The MG architecture further refines this by enabling effective multi-scale processing. Furthermore, the results underscore the inadequacy of models restricted to narrow frequency bands, reinforcing the need for a holistic, full-spectrum approach as embodied by \ourmethod{}.

\subsubsection{Qualitative Analysis of Vorticity Field Predictions}
\label{sec:qualitative_vorticity_analysis}

Beyond quantitative metrics, we qualitatively assessed the performance of the ablated \ourmethod{} variants by visualizing their predicted vorticity fields over an extended period, as depicted in Figure~\ref{fig:ablation_vorticity_visualization}. This visual inspection provides crucial insights into how well each model captures the intricate dynamics and preserves the physical realism of decaying turbulence.

The Ground Truth (top row) illustrates the complex evolution of turbulent eddies, characterized by stretching, merging, and filamentation, with a rich spectrum of scales. The `w/o MG` variant (bottom row), which retains the HDS mechanism, demonstrates a remarkable ability to track these complex dynamics. Its predicted fields maintain a high degree of structural similarity to the Ground Truth throughout the entire simulation up to $t=88$. Fine-scale vortex filaments and the overall morphology of larger eddies are well-preserved, indicating that the HDS mechanism effectively captures cross-scale interactions and mitigates excessive numerical dissipation or smoothing often seen in long-term autoregressive predictions.

In stark contrast, the `w/o HDS` variant (fourth row) completely fails to produce physically plausible results. From early time steps, its predictions deviate significantly, and by $t=22$ and beyond, the fields degenerate into non-physical, patchy structures, bearing no resemblance to the true turbulent flow. This visually confirms the findings from the loss and error metrics: the HDS mechanism is absolutely fundamental for learning the underlying physics.

The `HighFreq Only` (second row) and `LowFreq Only` (third row) variants also exhibit significant deficiencies. While the `HighFreq Only` model initially captures some smaller-scale features, it progressively loses coherence and energy, leading to an overly smoothed field by $t=55$. The `LowFreq Only` model, from the outset, fails to resolve fine-scale structures, resulting in a blurry and overly diffuse representation of the vorticity field. As time progresses, it loses almost all small-scale information, retaining only vague, large-scale outlines that deviate substantially from the Ground Truth. These observations underscore the necessity of a full-spectrum approach, as models focusing on limited frequency bands cannot sustain high-fidelity predictions of the multi-scale turbulent cascade.

These qualitative results strongly corroborate our quantitative findings and further highlight the efficacy of the HDS mechanism in \ourmethod{}. Even without the multi-grid framework, HDS enables robust and physically consistent long-term tracking of turbulence evolution, a feat that other ablated versions are unable to achieve. The superior performance of the full \ourmethod{} model would build upon this strong foundation laid by HDS, with the MG architecture further enhancing the resolution of fine-scale details and the accuracy of cross-scale energy transfer.


\subsection{Visualization of Long-Term Prediction on Forced Isotropic Turbulence}
\label{sec:viz_forced_turbulence}
To further demonstrate the robustness and superior long-term prediction capabilities of \ourmethod{}, we present a qualitative comparison against several baseline methods on the Forced Isotropic Turbulence benchmark. This benchmark involves continuously forced dynamics, posing a distinct challenge compared to decaying turbulence as the model must accurately capture both the externally driven evolution and the internal turbulent cascade. Figure~\ref{fig:forced_turbulence_visualization} displays the predicted vorticity fields from an initial condition up to $t=19$.

As illustrated, \ourmethod{} (Turb-L1, second row) consistently generates predictions that closely mirror the ground truth evolution (approximated by the first row's sequence after the initial condition). Even at later stages like $t=17$ and $t=19$, \ourmethod{} successfully preserves the sharpness of vortex edges, the intricate details of smaller eddies, and the overall complex topology of the turbulent flow. This high degree of visual fidelity indicates \ourmethod{}'s effectiveness in maintaining spectral accuracy across a wide range of wavenumbers and its ability to handle the non-linear interactions inherent in forced turbulence.

In contrast, all baseline methods exhibit noticeable degradation over time. Convolutional Neural Operators (CNO, third row) and PastNet (fourth row) begin to show signs of excessive smoothing and loss of fine-scale details from intermediate time steps (around $t=7$ to $t=11$). By $t=19$, their predictions, while retaining some large-scale features, lack the rich high-frequency content present in the ground truth and \ourmethod{}'s output. ResNet (fifth row) displays even more pronounced smoothing artifacts from earlier on, failing to capture the strength and definition of many vortical structures. SimVP (sixth row) also struggles with long-term fidelity, its predictions becoming progressively blurred and distorted, losing critical structural information.

These visual results on the forced turbulence scenario reinforce the conclusions drawn from the decaying turbulence experiments and quantitative metrics. The inherent spectral bias in many conventional deep learning architectures, including those designed for spatiotemporal forecasting or operator learning, limits their ability to perform accurate, high-fidelity, long-range predictions of complex fluid dynamics. \ourmethod{}, with its explicit mechanisms for hierarchical dynamics synthesis and overcoming spectral bias, demonstrates a clear advantage in robustly tracking the full evolution of turbulence under continuous forcing, making it a promising approach for practical scientific and engineering applications requiring reliable long-term forecasting.

\section{Conclusion}

In this paper, we present the first spectral bias-based analysis of the over-smoothing phenomenon in long-term turbulence prediction. Through extensive experiments, we demonstrate that existing models often overlook critical high-frequency details during training. Building on these findings, we propose \ourmethod, which incorporates a hierarchical dynamics synthesis mechanism within a multi-grid architecture, enabling robust and efficient long-term prediction. Extensive benchmark tests demonstrate that \ourmethod achieves exceptional long-term accuracy and fidelity, effectively overcoming spectral bias while preserving physical realism. We anticipate that \ourmethod will provide a novel perspective and a reliable framework for modeling complex dynamics in Earth systems.

\ifCLASSOPTIONcaptionsoff
  \newpage
\fi



%
\bibliographystyle{IEEEtran}
\bibliography{main}

\end{document}